\documentclass[twocolumn, journal]{IEEEtran}

\usepackage{amsmath,amsfonts,amssymb,amsthm,mathtools,amsfonts}
\usepackage{color}
\usepackage{algorithm}
\usepackage{algpseudocode}
\usepackage{array}
\usepackage{mathtools}

\usepackage{stfloats}
\usepackage{url}
\usepackage{verbatim}
\usepackage{graphicx}
\usepackage{cite}
\usepackage{amsmath}
\usepackage{bbm}
\usepackage{cite}
\usepackage{kotex}
\usepackage{subfigure}
\usepackage[dvipsnames]{xcolor}

\makeatletter
\def\equalsfill{$\m@th\mathord=\mkern-7mu
\cleaders\hbox{$\mathord=$}\hfill
\mkern-7mu\mathord=$}
\makeatother

\newtheorem{lemma}{Lemma}

\newtheorem{theorem}{Theorem}
\newtheorem{remark}{Remark}

\newtheorem{assumption}{Assumption}
\allowdisplaybreaks

\begin{document}

\title{Two-Stage Wireless Federated LoRA Fine-Tuning with Sparsified Orthogonal Updates}
 
\author{
	\IEEEauthorblockN{Bumjun Kim},~\IEEEmembership{Graduate Student Member,~IEEE},
	and
	\IEEEauthorblockN{Wan Choi},~\IEEEmembership{Fellow,~IEEE}
	\vspace{-0.1in}
	\thanks{B.~Kim and W.~Choi are with the Department of Electrical and Computer Engineering, Seoul National University (SNU), and the Institute of New Media and Communications, SNU, Seoul 08826, Korea. (e-mail: \{eithank96, wanchoi\}@snu.ac.kr) (\emph{Corresponding Authors: Wan Choi})}
		
}

\maketitle

\begin{abstract}
Federated fine-tuning with low-rank adaptation (LoRA) communicates only two low-rank matrices instead of the full model, but existing methods typically fix the LoRA rank in advance as a manually tuned hyperparameter. In wireless networks, however, the rank determines both adaptation capacity and uplink payload, while the deliverable payload varies with the fading channel. To address this coupling, we formulate wireless federated LoRA fine-tuning as a two-timescale design that separates the \emph{offline-optimized rank}, i.e., the rank of the shared LoRA structure selected before training from statistical channel information, from the per-iteration sparsification and bandwidth decisions adapted to instantaneous CSI. For per-iteration adaptation, we propose sparsified orthogonal fine-tuning (\textbf{SOFT}), which promotes near-orthogonality among rank components so that the product of the corresponding column and row norms approximates each component's singular value. This SVD-free score guides component-wise payload allocation and within-component entry selection without forming the full matrix product. We further derive a convergence bound linking the two timescales and develop a two-stage federated algorithm (\textbf{TSFA}) that selects the offline-optimized rank offline and jointly optimizes sparsification and bandwidth online via Lyapunov optimization.
\end{abstract}

\begin{IEEEkeywords}
Parameter-efficient fine-tuning, federated learning, sparsification, Lyapunov optimization, low-rank adaptation.
\end{IEEEkeywords}
\IEEEpeerreviewmaketitle

\section{Introduction}
Large pre-trained transformer models have demonstrated exceptional performance across a wide range of tasks. However, adapting these models to specific downstream tasks typically requires fine-tuning, and updating the entire parameter set of a large model is computationally and memory intensive. Parameter-efficient fine-tuning (PEFT) methods \cite{houlsby2019parameter,hu2022lora} alleviate this burden by updating only a small subset of model parameters. Among them, low-rank adaptation (LoRA) \cite{hu2022lora} freezes the pre-trained weights and trains only a pair of low-rank matrices for each adapted layer, achieving performance comparable to full fine-tuning with only a small fraction of trainable parameters. In many practical scenarios, however, the data required for adaptation are distributed across multiple clients and cannot be centralized due to privacy concerns, making conventional centralized fine-tuning infeasible. Federated learning (FL \cite{10081195,chen2020joint,kim2024privacy} addresses this challenge by enabling clients to collaboratively fine-tune a shared model through the exchange of model updates rather than raw data.

Accordingly, the integration of LoRA and FL has emerged as a practical approach to distributed fine-tuning. The authors in \cite{zhang2024towards} pioneered the integration of LoRA with federated averaging. Subsequent studies have advanced federated LoRA from various perspectives, including aggregation under heterogeneous data distributions \cite{yi2025fedalora}, privacy-preserving training through partial LoRA freezing \cite{sun2024improving}, data-driven initialization \cite{babakniya2023}, split deployment between clients and an edge server \cite{wang2025federated,chen2025privacy}, privacy-aware split configuration with joint resource allocation \cite{Chen2025}, and dropout-based regularization \cite{xie2026fedlodrop}. Despite these advances, LoRA rank selection has received limited attention, with the rank typically treated as a manually tuned hyperparameter.

This unresolved rank-selection problem becomes particularly consequential when client updates must be transmitted over a wireless uplink. The available bandwidth is shared among scheduled clients, channel conditions vary across communication rounds, and the server cannot complete an aggregation round until all scheduled updates have been received \cite{huh2025feature,huh2026federated}. In this setting, the LoRA rank plays a dual role: 1) as a model parameter, it determines the adaptation capacity; and 2) as a communication parameter, it determines the uplink payload in each communication round because the number of transmitted parameters scales linearly with the rank. These two roles directly conflict under a latency constraint: increasing the rank can improve adaptation capacity but also increases the communication load. A principled wireless federated LoRA design must therefore treat the rank jointly as a learning and communication variable rather than solely as a learning hyperparameter.

Existing rank-selection strategies do not fully address this requirement. In federated LoRA, the server instantiates a common LoRA structure before training begins; we refer to the rank of this structure as the \emph{offline-optimized rank}. The offline-optimized rank defines the parameter space in which client updates are aggregated and determines the maximum uplink payload of each client. Because conventional federated aggregation directly averages the two LoRA matrices, all participating clients must share a common matrix structure. Consequently, the offline-optimized rank must be fixed before training and remain unchanged throughout training. Existing studies, however, typically choose this rank without explicitly accounting for either the learning task or the wireless channel, even though an inappropriate choice can be costly in both respects. A rank that is too small may provide insufficient adaptation capacity, whereas a rank that is too large may generate an uplink payload that cannot be delivered within the latency constraint. Adaptive-rank methods \cite{cho2023heterogeneous,wang2024flora} do not resolve this issue either. These methods assign each client a subset of the rank components within a predefined LoRA structure, allowing the number of active components to vary across clients or over training while taking the rank of the underlying structure as given. In other words, existing methods adapt how a predetermined offline-optimized rank is used rather than determining what the offline-optimized rank itself should be. We address precisely this problem by selecting the offline-optimized rank before training according to the learning and channel conditions under which federated fine-tuning will operate.

Even after the offline-optimized rank is selected, a second decision remains at every iteration. The offline-optimized rank, chosen before training from statistical channel knowledge, determines how rich an update the model can express. However, how much of that update can actually be delivered within the latency constraint varies with the instantaneous state of the fading channel, and each client must therefore choose which part of its update to send. Recall that a LoRA update is represented as the product of two matrices and can be decomposed into rank components, each formed by a column of the first matrix paired with the corresponding row of the second. Selecting what to transmit therefore requires a criterion for identifying the most important components, yet existing compression strategies are poorly suited to this task.

Existing methods perform selection either entry-wise or in a structured manner. Entry-wise schemes apply generic top-$q$ selection directly to the LoRA matrices as if they were independent, unstructured parameter blocks \cite{stich2018sparsified,kuo2024federated,gao2025federated}. Such approaches overlook the multiplicative coupling between the two matrices, because the contribution of each rank component is jointly determined by its corresponding column and row. Structured schemes instead operate on entire columns and rows, retaining a prescribed subset of rank components according to their indices \cite{cho2023heterogeneous} or randomly zeroing rows and columns of the LoRA matrices \cite{xie2026fedlodrop}. In both cases, the selection does not explicitly account for the importance of the retained components. Exactly identifying the most influential components would require forming the full LoRA update matrix and computing its singular value decomposition (SVD) at every iteration, which is prohibitively expensive for resource-constrained clients. Furthermore, because channel conditions vary across iterations, the transmitted payload and bandwidth allocation should be jointly adapted to the instantaneous channel state. Wireless federated LoRA therefore requires both a computationally efficient importance criterion and a per-iteration control mechanism that adapts the transmitted payload to the wireless channel.

We address these challenges through a two-timescale design, following a principle widely used in wireless resource allocation in which slowly varying decisions are optimized using statistical channel state information (CSI), while rapidly varying decisions are adapted using instantaneous CSI. The offline-optimized rank naturally belongs to the slow timescale: it is determined before training, when only statistical CSI is available, and remains fixed throughout the training process. In contrast, the sparsification ratio and bandwidth allocation belong to the fast timescale because they can be adjusted at every iteration after the instantaneous CSI is observed. Guided by this separation, we propose a two-stage federated algorithm (\textbf{TSFA}). In the offline stage, TSFA selects the offline-optimized rank by minimizing a convergence bound that captures the effects of rank on both learning performance and communication latency under statistical CSI. In the online stage, TSFA jointly controls the sparsification ratio and bandwidth allocation through Lyapunov optimization using only the instantaneous CSI of the current iteration, while provably satisfying the time-average latency constraint.

To enable importance-aware sparsification, we further propose sparsified orthogonal fine-tuning (\textbf{SOFT}). An orthogonal regularizer incorporated into the training loss encourages the rank components of the LoRA matrices to become nearly orthogonal. Under this structure, the product of the column and row norms associated with each rank component approximates its singular value. \textbf{SOFT} therefore enables SVD-guided importance estimation without explicitly forming the full product matrix or performing an SVD, both of which are prohibitively expensive for resource-constrained clients. At each iteration, the available transmission budget is distributed across rank components in proportion to their importance scores. Components with negligible scores receive few or no entries and are therefore effectively deactivated, inducing sparsification along the rank dimension. Within each retained component, only the largest-magnitude entries are selected according to its allocated budget, thereby additionally inducing entry-wise sparsification.
  
\textbf{Contributions.}\quad The key contributions are as follows.
\begin{itemize}

\item We formulate wireless federated LoRA fine-tuning as a two-timescale design that separates the \emph{offline-optimized rank}, fixed before training, from the per-iteration sparsification and bandwidth decisions adapted to instantaneous CSI. This separation assigns adaptation capacity to the slow timescale and channel-dependent communication payload to the fast timescale.
    
\item We develop \textbf{SOFT}, which promotes near-orthogonality among rank components and uses a per-rank norm-based score as an SVD-free importance proxy. This enables importance-aware sparsification without forming the full LoRA product matrix or performing repeated SVDs.
    
 \item We derive a convergence bound that captures offline-optimized-rank approximation error, partial client participation, sparsification error, and separate-matrix aggregation discrepancy. The deived bound reveals the tradeoff among adaptation capacity, communication load, and optimization accuracy. We also empirically validate the underlying spectral-decay and covariance assumptions.
    
\item Based on this analysis, we develop \textbf{TSFA}, which selects the offline-optimized rank offline using statistical CSI and jointly adapts the sparsification ratio and bandwidth online using instantaneous CSI. Experiments show higher accuracy than existing baselines under the same latency constraint.
\end{itemize}

\section{System Model}\label{sec: system model}
\begin{table}[t!]
\caption{Notations and Descriptions}
\vspace{-1mm}
\label{tab:Notations}
\footnotesize
\centering
\begin{tabular}{cl}
\hline
\textbf{Notation}                                    & \textbf{Description}                                                                                     \\ \hline
$\theta_P$ & Pre-trained model weight matrix in $\mathbb{R}^{d \times \ell}$ \\
$\mathcal{D}_k$ & Local dataset of client $k$ \\
$B$ & Total available bandwidth \\
$b_k^t$ & Bandwidth allocation ratio for client $k$ at iteration $t$ \\
$h_k^t$ & Channel gain for client $k$ at iteration $t$ \\
$D^t$ & Overall transmission delay at iteration $t$ \\
$\bar{D}$ & Upper bound of $D^t$\\
$\mathcal{K}^t$ & Set of scheduled clients at iteration $t$ \\
$m_{k}^t$ & Error feedback memory matrix at client $k$ \\
$\tilde{m}_k^t$ & Concatenated error feedback vector \\
$O^t$ & Sparsification ratio at iteration $t$ \\
$\phi$ & Constant used in the covariance bound assumption \\
$S$ & Smoothness constant of the loss functions \\
$G_k$ & Upper bound on the gradient norm at client $k$ \\
$G^2$ & Maximum gradient bound, $G^2 = \max_{k}G_k^2$ \\
$Q^t$ & Virtual queue at iteration $t$ \\
$D_{th}$ & Average transmission delay threshold\\ \hline
\end{tabular}
\end{table}
\subsection{Low-rank Adaptation}\label{subsec:LoRA}
Suppose we are given a pre-trained model $\theta_P \in \mathbb{R}^{d \times \ell}$ to be adapted to a downstream task. In conventional fine-tuning, the model is initialized with $\theta_P$ and updated via gradient descent as $\theta = \theta_P + \Delta\theta$, where $\Delta\theta$ contains updates for all $d \times \ell$ parameters. 

Rather than directly optimizing $\Delta\theta$, LoRA represents the weight update using the low-rank decomposition $\Delta\theta = \theta_B \theta_A$, where $\theta_B \in \mathbb{R}^{d \times r}$, $\theta_A \in \mathbb{R}^{r \times \ell}$, and $r \ll \min(d, \ell)$. The resulting output is
\begin{align}
    \boldsymbol{y}^\textsf{T}=\boldsymbol{x}^\textsf{T}\theta_P +\frac{\alpha}{r}\boldsymbol{x}^\textsf{T}\Delta\theta =\boldsymbol{x}^\textsf{T}\theta_P +\frac{\alpha}{r} \boldsymbol{x}^\textsf{T}\theta_B\theta_A,
\end{align}
where $\boldsymbol{y} \in \mathbb{R}^{\ell \times 1}$, $\boldsymbol{x} \in \mathbb{R}^{d \times 1}$, and $\frac{\alpha}{r}$ is a scaling factor. During fine-tuning, $\theta_P$ remains fixed and only $\theta_A$ and $\theta_B$ are updated. This reduces the number of trainable parameters by a factor of $\mathcal{O}\left( \frac{r}{\min(d, \ell)} \right)$ relative to full fine-tuning.

The offline-optimized rank $r$ directly controls the expressive capacity of the LoRA update and therefore affects fine-tuning performance. A larger rank can represent a richer set of weight updates and potentially improve adaptation accuracy. However, over-parameterized models often exhibit low intrinsic dimensionality~\cite{aghajanyan2020intrinsic,hu2022lora}, suggesting that a sufficiently large but finite rank can achieve performance comparable to full fine-tuning without requiring a full-rank update.

To preserve the pre-trained model at initialization, $\theta_B$ is initialized to zero, while $\theta_A$ is initialized with random Gaussian values. The scaling parameter $\alpha$ controls the magnitude of the LoRA update and helps stabilize training. By optimizing only the low-rank matrices, LoRA substantially reduces memory and computational requirements, making it well suited for fine-tuning large models on resource-constrained clients.

\subsection{Federated LoRA Fine-Tuning}\label{subsec:FLoRA}
Consider a wireless federated learning network consisting of a central parameter server and $N$ single-antenna clients that collaboratively fine-tune a global model $\theta\in\mathbb{R}^{d \times \ell}$ using a LoRA module with offline-optimized rank $r$. When LoRA is integrated with FL, the pre-trained model $\theta_P$ remains fixed at both the server and the clients and therefore does not need to be exchanged during training. Instead, only the LoRA matrices $\theta_B$ and $\theta_A$ are locally updated and communicated in each round, substantially reducing the communication overhead compared with full fine-tuning, which requires exchanging the entire model.

Each client $k\in\mathcal{N}$ owns a local dataset $\mathcal{D}_k=\{(x_i,y_i)\}_{i=1}^{D_k}$ where $D_k=|\mathcal{D}_k|$ and $\mathcal{D}=\cup_{k\in\mathcal{N}}\mathcal{D}_k$.  The local loss function of client $k$ is 
\begin{align}
    F_k(\theta_B,\theta_A)=\frac{1}{|\mathcal{D}_k|}\sum_{x_i\in\mathcal{D}_k}f(\theta_B,\theta_A;x_i,y_i),
\end{align}
where $f(\theta_B,\theta_A)$ denotes the  sample-wise loss function. Accordingly, the global loss function is
\begin{align}
    F(\theta_B,\theta_A)=\sum_{k\in\mathcal{N}}p_kF_k(\theta_B,\theta_A),
\end{align}
where $p_k=|\mathcal{D}_k|/|\mathcal{D}|$ and $\sum_{k\in\mathcal{N}}p_k=1$. The objective of federated LoRA fine-tuning is to obtain the optimal LoRA matrices by solving
\begin{align}
(\theta_B^\ast,\theta_A^\ast)
=\underset{\theta_B\in\mathbb{R}^{d \times r},\theta_A\in\mathbb{R}^{r \times \ell}}{\arg\min}
F(\theta_B,\theta_A),
\end{align}
which yields the fine-tuned global model $\theta^\ast=\theta_P+\frac{\alpha}{r}\theta_B^\ast\theta_A^\ast$.

The objective $F(\theta_B,\theta_A)$ is progressively minimized through alternating local updates and global aggregation. The procedure follows FedAvg, except that only the LoRA matrices are communicated and aggregated. Specifically, at communication round $t=1,2,\dots,T$, the following steps are performed:
\begin{enumerate}
\item \textbf{Broadcasting:} The server broadcasts the global LoRA matrices $\theta_B^{t}$ and $\theta_A^{t}$ to the scheduled client set $\mathcal{K}^t$. Since the server is assumed to have sufficient transmit power, the downlink transmission is considered error-free.

\item \textbf{Local Fine-Tuning:} Each scheduled client $k\in\mathcal{K}^t$ initializes its local LoRA matrices using the received global matrices, i.e., $\theta_{B,k}^{t}=\theta_B^{t}$ and $\theta_{A,k}^{t}=\theta_A^{t}$. The client then performs local fine-tuning using a gradient-based optimizer, such as gradient descent or Adam, to obtain the updated matrices $\theta_{B,k}^{t+1}$ and $\theta_{A,k}^{t+1}$.

\item \textbf{Matrix Aggregation:} Each scheduled client $k\in\mathcal{K}^t$ uploads $\theta_{B,k}^{t+1}$ and $\theta_{A,k}^{t+1}$ to the server. The server separately aggregates the two matrices to obtain the global LoRA parameters for the next round:
  \begin{align}        
    \theta_B^{t+1}=\frac{N}{K}\sum_{k\in\mathcal{K}^{t}}p_k\theta_{B,k}^{t+1}, \;\;\theta_A^{t+1}=\frac{N}{K}\sum_{k\in\mathcal{K}^{t}}p_k\theta_{A,k}^{t+1},
    \end{align}
where $K=|\mathcal{K}^t|$ denotes the number of scheduled clients.
\end{enumerate}
After $T$ communication rounds, the resulting global LoRA matrices define the fine-tuned model. In principle, each client could upload the complete LoRA update $\theta_{B,k}\theta_{A,k}$. However, explicitly forming this product incurs additional computation at resource-constrained clients and produces a $d\times\ell$ matrix, which is substantially larger than the two low-rank factors when $r\ll\min(d,\ell)$. Therefore, clients transmit $\theta_{B,k}$ and $\theta_{A,k}$ separately. This avoids local matrix multiplication while reducing the number of transmitted parameters from $d\ell$ to $r(d+\ell)$, thereby lowering both communication overhead and uplink latency.
 
\subsection{Communication Model}
We consider wireless communication between the local clients and the server. In practice, the uplink used to transmit LoRA updates is typically more bandwidth-constrained than the downlink. We therefore focus on uplink communication, while assuming that the downlink provides reliable transmission at effectively unlimited rates due to the server's higher transmit power \cite{kim2024privacy}.

For the uplink, we adopt a frequency-division multiple access (FDMA) scheme with total bandwidth $B$. The channel gain between client $k$ and the server at iteration $t$ is modeled as $h_k^t = z_k^t \delta_k^{-\chi}$, where $z_k^t \sim \mathcal{CN}(0,1)$ denotes the small-scale fading coefficient and $\delta_k^{-\chi}$ represents distance-dependent path loss with exponent $\chi$. Let $b_k^t \in [0,1]$ denote the bandwidth allocation ratio for client $k \in \mathcal{N}$, satisfying $\sum_{k=1}^{N} b_k^t=1$. The scheduling variable $a_k^t \in \{0,1\}$ indicates whether client $k$ is selected at iteration $t$, with $\sum_{k=1}^{N} a_k^t=K$. If $a_k^t=0$, no bandwidth is allocated to client $k$, i.e., $b_k^t=0$.
 
 The uplink transmission rate of client $k$ is given by
\begin{align}
j_k^t = b_k^t B \log_2\left(1 + \frac{|h_k^t|^2}{\sigma^2}\right),
\end{align}
where $\sigma^2$ denotes the variance of the additive white Gaussian noise (AWGN). Let $v$ denote the number of bits required per model parameter. The communication delay of client $k$ at iteration $t$ is then
\begin{align}\label{eq: delay}
    D_k^t = \frac{a_k^t vO^t r(d + \ell)}{j_k^t} = \frac{a_k^t v O^t r(d + \ell)}{b_k^t B \log_2\left(1 + \frac{|h_k^t|^2}{\sigma^2}\right)},
\end{align}
where $O^t$ denotes the sparsification ratio defined in Section \ref{sec: OLoRA}. If $a_k^t=0$, we set $D_k^t=0$. Since each communication round is completed only after all scheduled clients finish their uplink transmission, the overall delay at iteration $t$ is determined by the slowest scheduled client:
\begin{align}
D^t = \max_{k \in \mathcal{K}^t} D_k^t.
\end{align}

\section{Sparsified Orthogonal Fine-Tuning}\label{sec: OLoRA}

\subsection{Loss Function Design}
To reduce communication overhead in FL, model updates are commonly sparsified before transmission. An effective sparsification rule should therefore retain the most informative components of each update. Among all rank-$r$ approximations of a matrix, truncated SVD provides the optimal approximation in the Frobenius norm \cite{eckart1936approximation}, suggesting that dominant singular directions offer a principled measure of component importance. However, computing an SVD at every communication round is prohibitively expensive for resource-constrained clients in LoRA-based FL. We therefore seek a lightweight surrogate that captures this structural property without explicitly performing matrix decomposition.

A key component of our sparsification method is to encourage near-orthogonality in the LoRA matrices $\theta_{B,k}$ and $\theta_{A,k}$. Specifically, we promote $\theta_{B,k}^\textsf{T}\theta_{B,k}$ and $\theta_{A,k}\theta_{A,k}^\textsf{T}$ to be approximately diagonal, analogous to the orthogonality of the left and right singular vectors in SVD. This makes the columns of $\theta_{B,k}$ and the rows of $\theta_{A,k}$ nearly orthogonal and enables each rank component to be evaluated independently.

To impose this structure, client $k$ minimizes
\begin{align}\label{eq:loss_function}
    L_k=L_{task,k}&+\zeta(\|\theta_{B,k}^\textsf{T}\theta_{B,k}-diag(\theta_{B,k}^\textsf{T}\theta_{B,k})\|_F^2\nonumber\\
    &+\|\theta_{A,k}\theta_{A,k}^\textsf{T}-diag(\theta_{A,k}\theta_{A,k}^\textsf{T})\|_F^2),
\end{align}
where $L_{task,k}$ denotes the task-specific loss, $\zeta$ controls the strength of the orthogonal regularization, and $\operatorname{diag}(C)$ denotes the diagonal matrix formed from the diagonal entries of $C$. This regularization encourages the LoRA factors to exhibit SVD-like orthogonality without requiring explicit decomposition.

Under this near-orthogonal structure, the Frobenius norm of the LoRA update can be approximated as
\begin{align} 
\|\theta_{B,k}\theta_{A,k}\|_F^2 \approx \sum_{i=1}^r \|\theta_{B,k}[:,i]\|_2^2 \|\theta_{A,k}[i,:]\|_2^2, 
\end{align}
where $\theta_{B,k}[:,i]$ and $\theta_{A,k}[i,:]$ denote the $i$-th column and row, respectively. The cross-component terms become negligible due to near-orthogonality, allowing the contribution of each rank component to be quantified using only the product of its column and row norms.

\subsection{Sparsification Method}
At the beginning of each iteration, client $k$ performs local training using the proposed loss function \eqref{eq:loss_function}, obtaining the updated LoRA matrices $\theta_{B,k}$ and $\theta_{A,k}$. The client then sparsifies these matrices by retaining a total of $r(d+\ell)O$ elements, where $O\in[0,1]$ denotes the sparsification ratio, and setting the remaining elements to zero.

When the columns of $\theta_{B,k}$ and the rows of $\theta_{A,k}$ are nearly orthogonal, the product of their norms provides an approximation to the singular value associated with each rank component. Accordingly, we define the importance score of the $i$-th rank component as $s_{k,i} = \|\theta_{B,k}[:,i]\|_2\|\theta_{A,k}[i,:]\|_2$. Thus, $s_{k,i}^2$  approximates the corresponding squared singular value and serves as a measure of component importance. Let $o_{k,i}$ denote the number of nonzero elements allocated to the $i$-th rank component across $\theta_{B,k}[:,i]$ and $\theta_{A,k}[i,:]$. We distribute the total sparsification budget across rank components in proportion to their importance scores as
\begin{align}\label{eq:o_k,i}
    o_{k,i}=\frac{O r(d+\ell)\|\theta_{B,k}[:,i]\|_2^2\|\theta_{A,k}[i,:]\|_2^2}{\sum_{i=1}^r\|\theta_{B,k}[:,i]\|_2^2\|\theta_{A,k}[i,:]\|_2^2}.
\end{align}

\begin{figure}[!t]
    \centering
    \setlength{\subfigcapskip}{-2mm}
    \subfigure[]{\includegraphics[width=0.48\linewidth]{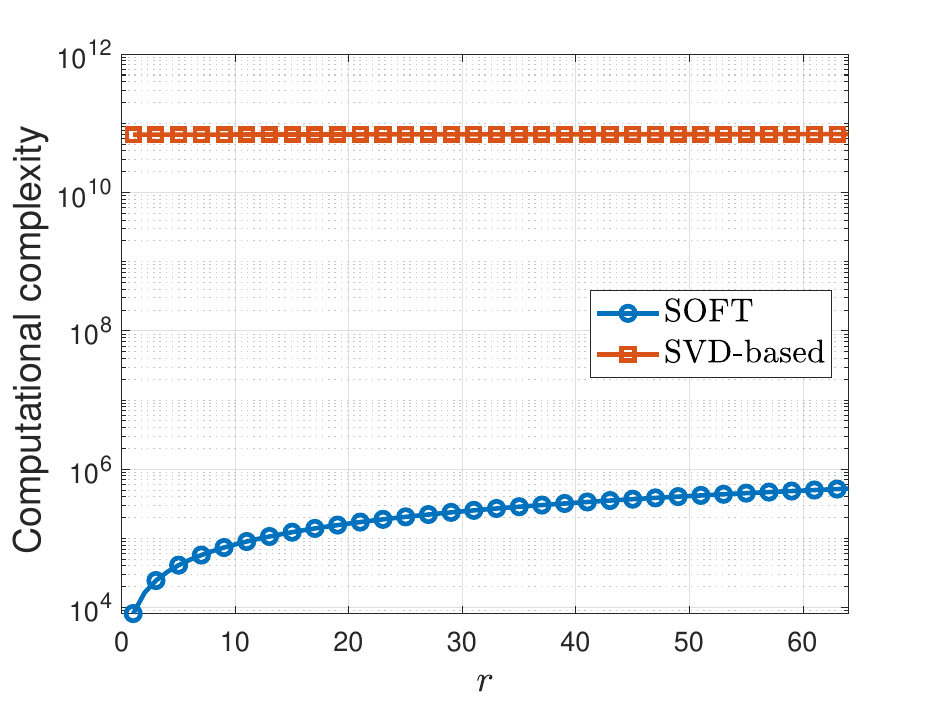}\label{fig: complexity1}}
    \subfigure[]{\includegraphics[width=0.48\linewidth]{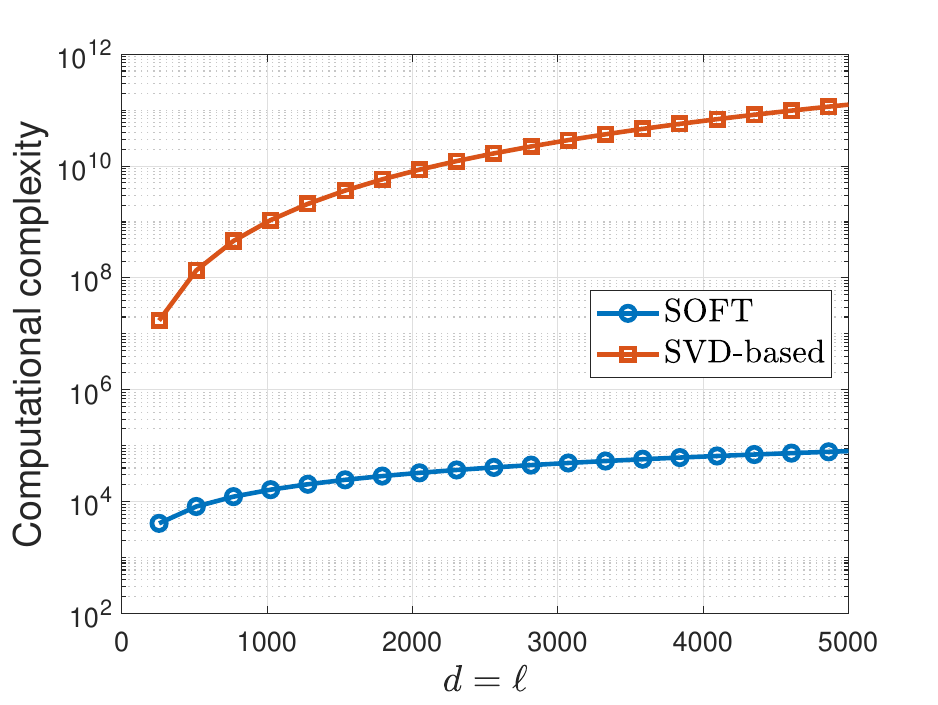}\label{fig: complexity2}}
    \caption{Computational complexity comparison between \textbf{SOFT} and the conventional full-multiplication and SVD-based approach. (a) Complexity versus the LoRA rank $r$. (b) Complexity versus matrix dimension.}
    \label{fig: complexity}
\end{figure}

In implementation, $o_{k,i}$ is converted to an integer allocation while preserving the total budget $Or(d+\ell)$. Note that  \eqref{eq:o_k,i} allocates more retained elements to rank components with larger measured importance. Consequently, components with negligible scores receive few or no elements and are effectively deactivated. The resulting sparsification therefore operates both across the rank dimension and within each retained component. Based on the allocation $o_{k,i}$, client $k$ retains the largest-magnitude entries associated with each rank component within its allocated budget.

The proposed sparsification requires only the computation of the $\ell_2$ norms of $r$ column-row pairs in each iteration, resulting in a complexity of $\mathcal{O}(r(d+\ell))$. This overhead is negligible compared with explicitly forming the full LoRA update and performing an SVD, whose complexity is $\mathcal{O}(dr\ell+\min(d,\ell)\cdot d\ell)$. To further demonstrate the computational efficiency of \textbf{SOFT}, Fig.~\ref{fig: complexity} compares its complexity with that of the conventional full-multiplication and SVD-based approach for different LoRA ranks and matrix dimensions. In Fig.~\ref{fig: complexity1}, we fix $d=\ell=4096$ and vary the LoRA rank $r$, whereas in Fig.~\ref{fig: complexity2}, we fix $r=8$ and vary the matrix dimension with $d=\ell$. The $y$-axis is shown on a logarithmic scale due to the substantial complexity gap between the two approaches. The results confirm that the additional computational overhead introduced by \textbf{SOFT} is negligible compared with explicit matrix multiplication and SVD.

\begin{figure}[!t]
    \centering
    \setlength{\subfigcapskip}{-2mm}
    \subfigure[]{\includegraphics[width=0.48\linewidth]{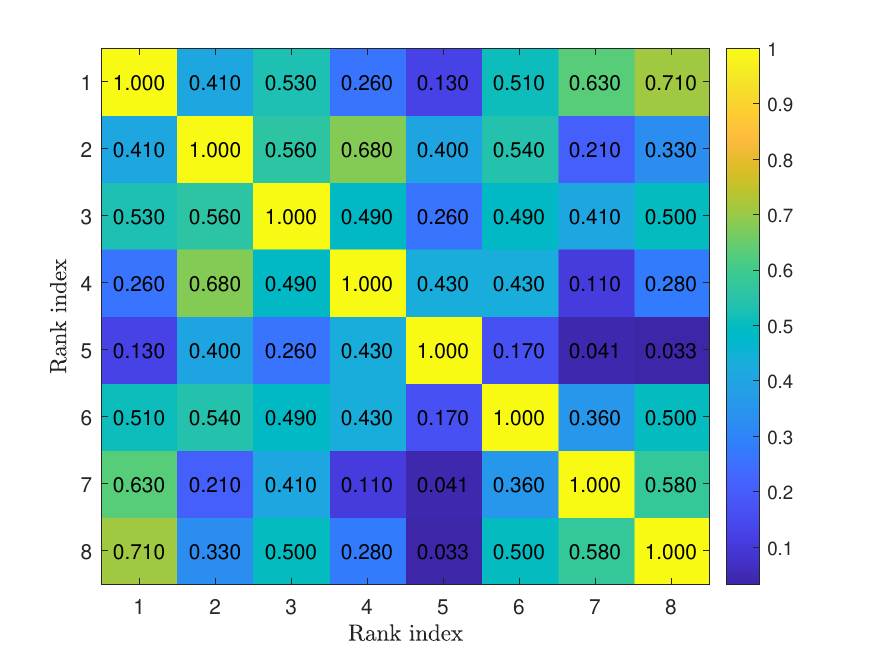}
    \label{fig: cov1}}
    \subfigure[]{\includegraphics[width=0.48\linewidth]{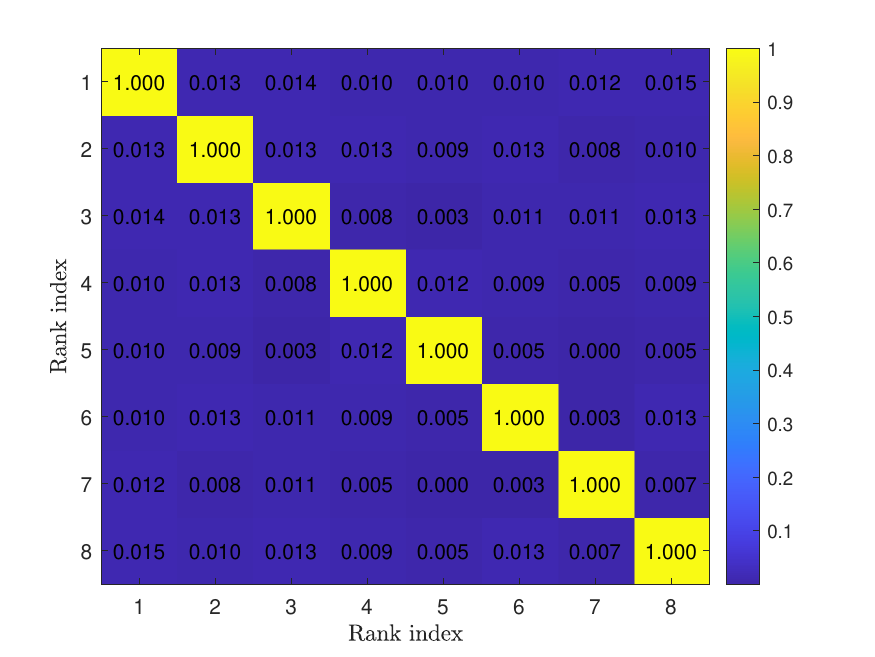}
    \label{fig: cov2}}
    \caption{$\theta_{B,k}^\textsf{T}\theta_{B,k}$ of Vit-Base blocks.6.mlp.fc1 layer: (a) without and (b) with the proposed loss function.}
\end{figure}

\begin{figure}[!t]
    \centering
    \includegraphics[width=0.8\linewidth]{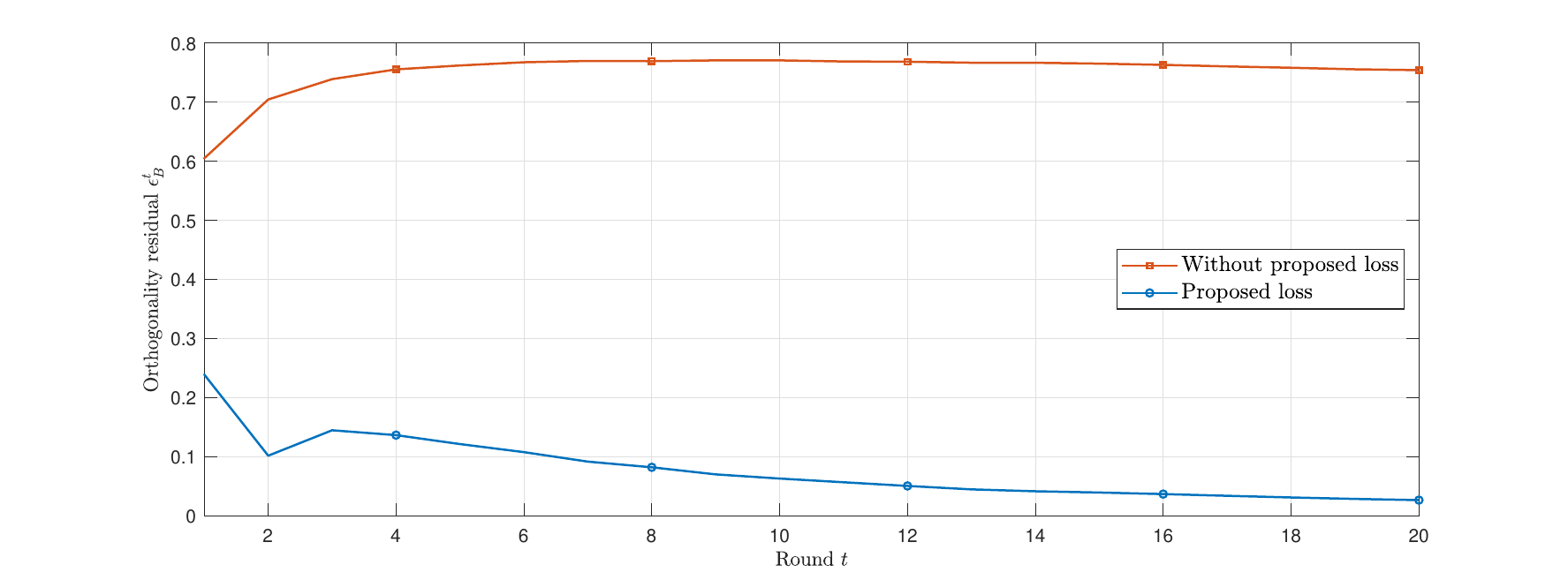}
    \caption{Normalized orthogonality residual $\epsilon_B^t$ of the
    same layer over training, without and with the proposed loss
    function.}
    \label{fig: eps_curve}
\end{figure}

 Figs.~\ref{fig: cov1} and \ref{fig: cov2} illustrate the effect of the proposed orthogonal regularization. After training with the proposed loss, the normalized Gram matrix $\theta_{B,k}^{\mathsf{T}}\theta_{B,k}$ becomes approximately diagonal, indicating increased orthogonality among the columns of $\theta_{B,k}$. Fig.~\ref{fig: eps_curve} further quantifies this behavior using the normalized orthogonality residual $\epsilon_B^t = \|\theta_{B,k}^{t\mathsf{T}}\theta_{B,k}^{t} - \mathrm{diag}(\theta_{B,k}^{t\mathsf{T}}\theta_{B,k}^{t})\|_F / \|\theta_{B,k}^{t\mathsf{T}}\theta_{B,k}^{t}\|_F$. Under the proposed loss, $\epsilon_B^t$ steadily decreases, whereas it remains nearly unchanged without the orthogonal regularization.


\begin{remark}
When the loss function  $\eqref{eq:loss_function}$ drives $\theta_B^\textsf{T}\theta_B$ and $\theta_A\theta_A^\textsf{T}$ toward diagonal matrices, the cross-component terms in $|\theta_B\theta_A|_F^2$ become negligible. Hence,
$\|\theta_B\theta_A\|_F^2=\sum_{i=1}^r \|\theta_B[:,i]\|_2^2\|\theta_A[i,:]\|_2^2,$
and the per-rank score $\|\theta_B[:,i]\|_2^2\|\theta_A[i,:]\|_2^2$ serves as a surrogate for the corresponding squared singular value. Therefore, allocating nonzero entries according to $o_{k,i}\propto \|\theta_B[:,i]\|_2^2\|\theta_A[i,:]\|_2^2$ provides an SVD-guided sparsification rule. As shown in Fig.~\ref{fig: eps_curve}, the decreasing residual $\epsilon_B^t$ supports the near-orthogonality induced by the proposed loss, enabling $s_{k,i}^2$ to closely track the squared singular values without explicitly forming the matrix product or performing an SVD.
\end{remark}

\subsection{Error Feedback with \textbf{SOFT}}
While the proposed sparsification method reduces communication overhead by transmitting only the most significant elements of $\theta_{B,k}$ and $\theta_{A,k}$, sparsification inevitably introduces approximation errors. To mitigate their impact on convergence, we incorporate an error-feedback mechanism into \textbf{SOFT} \cite{stich2018sparsified}. Each client maintains local error memories $m_{B,k}$ and $m_{A,k}$ that accumulate the information discarded by previous sparsification operations. Before sparsification, these residuals are added back to the current local matrices, yielding
\begin{align}
    \psi_{B,k}^t=\mathcal{S}(m_{B,k}^t+\theta_{B,k}^t),
    {\psi}_{A,k}^t=\mathcal{S}(m_{A,k}^t+\theta_{A,k}^t),
\end{align}
where $\mathcal{S}$ denotes the proposed sparsification operator, and $\theta_{B,k}^t$ and $\theta_{A,k}^t$ are the local LoRA matrices obtained using the loss function \eqref{eq:loss_function}. The residuals discarded by sparsification are then accumulated as
\begin{align}
m_{B,k}^{t+1} = m_{B,k}^t +\theta_{B,k}^t-\psi_{B,k}^t,\label{eq: memory update of B}\\
m_{A,k}^{t+1} = m_{A,k}^t +\theta_{A,k}^t-\psi_{A,k}^t. \label{eq: memory update of A}
\end{align}
Thus, information removed in the current iteration is retained locally and incorporated into subsequent updates rather than being permanently discarded.

For compact notation, we define the concatenated matrices
 \begin{align}
&\tilde{m}_k^t = \begin{bmatrix} {m_{B,k}^{t\textsf{T}}}  m_{A,k}^t \end{bmatrix}, \nonumber\\
&\tilde{\theta}_k^t = \begin{bmatrix} {\theta_{B,k}^{t\textsf{T}}}  \theta_{A,k}^t \end{bmatrix}, \quad
\tilde{\psi}_k^t = \begin{bmatrix} {\psi_{B,k}^{t\textsf{T}}}  \psi_{A,k}^t \end{bmatrix},
\end{align}
where $[\cdot]$ denotes concatenation. The error-memory update can then be written compactly as
\begin{align}\label{eq: overall memory update}
    \tilde{m}_{k}^{t+1} = \tilde{m}_{k}^t +\tilde{\theta}_{k}^t-\tilde{\psi}_{k}^t.
\end{align}
This mechanism progressively compensates for information lost through sparsification and thereby limits the accumulation of compression error during federated training. The effectiveness of \textbf{SOFT} with error feedback is evaluated in Subsection \ref{sec: S-OLoRA}. We next characterize the error induced by \textbf{SOFT} under the following standard contraction assumption \cite{stich2018sparsified}.

\begin{assumption}[\!\!\cite{stich2018sparsified}]\label{as: sparsification}
    Let $\mathcal{S} : \mathbb{R}^{q} \rightarrow \mathbb{R}^{q}$ denote the proposed sparsification operator, and let $u$ be the number of retained nonzero elements, with $0 < u \leq q$. Then, for all $x \in \mathbb{R}^{q}$,
\begin{align}
|\mathcal{S}(x)-x|_2^2\leq\left(1-\frac{u}{q}\right)|x|_2^2.
\end{align}
\end{assumption}

\begin{lemma}\label{lemma:error_bound}
Suppose that $O^i\in[O_{\min},1]$ for $0\leq i\leq t$.
Under Assumption~\ref{as: sparsification}, the error memory satisfies,
\begin{align}
\|\tilde m_k^{t+1}\|_F^2&\leq\left(1-\frac{O^t}{2}\right)\|\tilde m_k^t\|_F^2+2\rho^{t}\|\tilde\theta_k^t\|_F^2,\label{eq:one_step_memory}
\\
\|\tilde m_k^{t+1}\|_F^2&\leq4\rho_{\max}\max_{0\leq i\leq t}\|\tilde\theta_k^i\|_F^2,\label{eq:uniform_memory}
\end{align}
where
$\rho^{t}=\frac{1-O^t}{O^t}$ and $\rho_{\max}=\frac{1-O_{\min}}{O_{\min}^2}.$
\end{lemma}
\begin{IEEEproof}
    See \emph{Appendix \ref{apdx of lemma:error_bound}}.
\end{IEEEproof}

\section{Two-stage Federated Fine-Tuning}\label{sec: WFLoRA}

\subsection{Convergence Analysis and Problem Formulation}
In LoRA-based FL, unlike conventional FL, the use of low-rank adaptation introduces two additional factors that must be explicitly accounted for in the convergence analysis. First, as discussed in Subsection \ref{subsec:LoRA}, the fine-tuning performance of LoRA depends strongly on the selected rank $r$. We therefore characterize the approximation error induced by restricting the model update to rank $r$. Following \cite{zeng2023expressive}, we introduce the following assumption.

\begin{assumption}\label{as: lora error}
Let $\theta_o$ denote a LoRA model with sufficient rank to achieve performance comparable to full fine-tuning, and let $\theta_r$ denote the LoRA model of rank $r$ trained using the loss function \eqref{eq:loss_function}. For the layers equipped with LoRA modules, there exist constants $C>0$ and $\beta>1/2$ such that
\begin{align}\label{eq:lora_error_bound}
    \mathbb{E}\big[\|\theta_o-\theta_r\|_F^2\big]
    \;\leq\; \frac{C^2}{2\beta-1}\, r^{1-2\beta}.
\end{align}
\end{assumption}
\begin{remark}
\textbf{Assumption~\ref{as: lora error}} is motivated by the spectral structure of low-rank approximation. Since $\theta_r$ is trained to approximate $\theta_o$ under a rank-$r$ constraint, its approximation error is expected to follow the discarded spectral energy $\sum_{i>r}\sigma_i^2(\theta_o)$. By the Eckart--Young theorem \cite{eckart1936approximation}, this quantity corresponds to the minimum rank-$r$ approximation error in the Frobenius norm. Fine-tuning updates of pre-trained models often concentrate their energy in a small number of dominant singular directions \cite{hu2022lora,aghajanyan2020intrinsic}. We therefore model the spectrum by the power-law envelope $\sigma_i(\theta_o)\leq C i^{-\beta}$. Comparing the spectral tail with $\int_{r}^{\infty}x^{-2\beta}\,dx$ gives
$\sum_{i>r}\sigma_i^2(\theta_o)\leq\frac{C^2}{2\beta-1}r^{1-2\beta}$,
where $\beta>1/2$ ensures convergence of the tail. Empirical evidence supporting this relationship is provided in Subsection~\ref{sec: validation}.
\end{remark}
Second, as discussed in Subsection \ref{subsec:FLoRA}, clients transmit the two LoRA matrices $\theta_B$ and $\theta_A$ separately rather than forming and transmitting the full product $\theta_B\theta_A$. Although this substantially reduces computation and communication overhead, separate aggregation introduces an additional discrepancy at the server \cite{wang2024flora,bai2024federated}. Specifically,
$\mathbb{E}_k[\theta_{B,k}\theta_{A,k}]\ne\mathbb{E}_k[\theta_{B,k}]\mathbb{E}_k[\theta_{A,k}]$. 
Letting $\theta_B=\mathbb{E}_k[\theta_{B,k}]$ and $\theta_A=\mathbb{E}_k[\theta_{A,k}]$, this discrepancy can be expressed as
\begin{align}\label{eq: covariance}
    \mathbb{E}_k&[\theta_{B,k}\theta_{A,k}]-\theta_B\theta_A\nonumber\\
    &=\mathbb{E}_k[(\theta_{B,k}-\theta_B)(\theta_{A,k}-\theta_A)]\\
    &=\sum_{i=1}^{r}\mathbb{E}\left[\left(\theta_{B,k}[:,i]-\theta_B[:,i]\right)\left(\theta_{A,k}[i,:]-\theta_A[i,:]\right)\right]\\
    &=\textit{Cov}(\theta_B,\theta_A).
\end{align}
The term $\textit{Cov}(\theta_B,\theta_A)$ therefore quantifies the discrepancy induced by separately aggregating the two LoRA factors. Because all clients begin each round from a common global model, stronger data heterogeneity causes their locally updated parameters to diverge in different directions \cite{li2019convergence}, which can increase this covariance term. Its magnitude also depends on the scale of the LoRA parameters, represented by $|\tilde{\theta}_r|_F^2$, where $\tilde{\theta}_r=[\theta_B^{\textsf{T}}\ \theta_A]$. We capture this behavior through the following assumption.
\begin{assumption}\label{as: covariance}
    Let $\theta_r$ denote the LoRA model with rank $r$. The expected Frobenius norm of the covariance of $\theta_B$ and $\theta_A$ satisfies 
    $\mathbb{E}[\|\textit{Cov}(\theta_B,\theta_A)\|_F^2] \leq \phi \mathbb{E}\|\tilde{\theta}_r\|_F^2,$
    where $\phi$ is a constant that quantifies the degree of data heterogeneity across the clients.
\end{assumption} 

\begin{remark} 
    \textbf{Assumption \ref{as: covariance}} captures the dependence of the separate-aggregation discrepancy on both client heterogeneity and the magnitude of the LoRA parameters. Empirical evidence supporting this relationship is provided in Subsection \ref{sec: cov}.
\end{remark}
Next, for theoretical analysis, we make the following assumptions typically made in analyzing FL family \cite{li2019convergence}.

\begin{assumption}\label{as: smooth}
    The local objective function $F_k(\theta)$ is $S$-smooth, i.e. $ \|\nabla F_k(\theta)- \nabla F_k(\theta')\|_F\leq S\|\theta-\theta'\|_F,$
    for all $\theta$. Consequently, $F_k(\theta')\leq F_k(\theta)+\nabla F_k(\theta)^\textsf{T}(\theta'-\theta)+\frac{S}{2}\|\theta'-\theta\|_F^2,$
    Since the global objective $F(\theta)$ is a weighted average of the local objectives, it is also $S$-smooth. Furthermore, $F(\theta)$ is lower bounded by $F(\theta^*)$.
    \end{assumption}
\begin{assumption}\label{as:local gradient unbiasedness}
The stochastic gradient at each client is unbiased: $\mathbb{E}[\nabla F_k^{\scriptstyle{t, e}}(\theta)] = \nabla F_k(\theta)$ for all $k$, $t$, and $e$.
\end{assumption}

\begin{assumption}\label{as: norm boundedness}
The expected squared norm of the stochastic gradient at client $k$ is bounded as
$\mathbb{E}[\|\nabla F_k^{\scriptstyle(t, e)}(\theta)\|_F^2] \leq G_k^2$.
Moreover, each rank component of the locally trained LoRA matrices satisfies
 $\|\theta_{B,k}^t[:,i]\|_2^2 \leq W^2$ and
$\|\theta_{A,k}^t[i,:]\|_2^2 \leq W^2$ for all $i\in\{1,\dots,r\}$.
\end{assumption}
\begin{theorem}\label{th: convergence}
    Based on \textbf{Assumptions \ref{as: sparsification}, \ref{as: lora error}, \ref{as: covariance}, \ref{as: smooth}, \ref{as:local gradient unbiasedness}, \ref{as: norm boundedness}}  and \textbf{Lemmas \ref{lemma:error_bound} and \ref{lemma: bound of (a2)}}, the gradient norm of LoRA-based FL after $T$ global iterations is upper-bounded as
    \begin{align}
        &\frac{1}{T} \sum_{t = 0}^{T - 1} \mathbb{E}\left[ \left\| \nabla F_r(\theta_r^t) \right\|_F^2 \right] \leq \frac{4}{\eta T} \left( \mathbb{E}\left[ F_r(\theta_r^{0})  - F^*\right] \right)\nonumber\\
        &+ \frac{1}{T} \sum_{t = 0}^{T - 1}\bigg(\frac{4S^2C^2}{2\beta-1}\,r^{1-2\beta} +\frac{32(N-K)}{K(N-1)}\eta^2S^2E^2G^2 \nonumber\\
        &+2\eta SE^2G^2
        +\frac{256\,N r^2S^2W^4}{K}\left(\rho^t+4\rho_{\max}\right)\left(3+16\rho_{\max}\right)\nonumber\\
        &+32S^2\phi rW^2+ \frac{E(E-1)(2E-1)S^2\eta^2}{3}G^2\bigg)=\gamma^T,
\end{align}
    where $\rho^t=\frac{1-O^t}{O^t}$; $\rho_{\max}=\frac{1-O_{\min}}{O_{\min}^2}$; $G^2=\max_k G_k^2$; $E$ and $\eta$ represent the number of local iterations and learning rate, respectively.
\end{theorem}
\begin{IEEEproof}
    See \emph{Appendix \ref{apdx of th: convergence}}.
\end{IEEEproof}

\begin{remark} \textbf{Theorem \ref{th: convergence}} reveals how the LoRA rank $r$ and sparsification ratio $O^t$ jointly govern the convergence--communication tradeoff. Increasing $r$ reduces the rank-approximation term $r^{1-2\beta}$, thereby improving model expressiveness, but simultaneously enlarges the sparsification- and aggregation-related terms and increases the communication payload. Conversely, reducing $O^t$ lowers the transmitted payload but increases $\rho^t$ and hence the sparsification error. The theorem therefore provides an explicit analytical basis for jointly designing the LoRA rank and communication variables.
\end{remark}

Building on \textbf{Theorem \ref{th: convergence}}, we aim to minimize the convergence upper bound by jointly determining the offline-optimized rank $r$, sparsification ratios $O^t$, and bandwidth allocations $b_k^t$, subject to communication and latency constraints. The resulting optimization problem is
\begin{subequations}\label{OP:P1}
\begin{align}
    \mathcal{P}_1 &\min_{r, O^t, b_k^t} \gamma^T,\\
    \text{s.t.} & \ \  r \leq r_{\max}, r\in\mathbb{Z}^+ \label{OP:P1 rank}\\
    & \ \  \frac{1}{T} \sum\nolimits_{t = 0}^{T - 1} D^t \leq D_{th}, \label{OP:P1 latency}\\
    & \ \  O_{\min} \leq O^t \leq 1, \label{OP:P1 sparsification}\\
    & \ \ 0\leq b_k^t \leq 1, \quad\sum\nolimits_{k=1}^{N} b_k^t=1, \label{OP:P1 bandwidth}
\end{align}
\end{subequations}
where \eqref{OP:P1 rank} constrains the feasible offline-optimized rank, \eqref{OP:P1 latency}  imposes the average uplink latency constraint,\footnote{In this paper, the delay constraint refers to the latency required to upload a client's model update to the server in each iteration.} \eqref{OP:P1 sparsification}  constrains the sparsification ratio with minimum value $O_{\min}$, and \eqref{OP:P1 bandwidth} specifies the feasible bandwidth allocation.

Problem $\mathcal{P}_1$ presents two main challenges. First, the offline-optimized rank $r$ must be determined before training because it fixes the shared LoRA structure and affects both adaptation capacity and communication payload. Hence, unlike $O^t$ and $b_k^t$, it cannot be treated as an online control variable. Although prior works \cite{cho2023heterogeneous,wang2024flora,gao2025federated} consider adaptive-rank strategies, they operate within a predefined offline-optimized-rank budget and therefore do not address the offline-optimized-rank selection problem considered here. Moreover, selecting $r$ online would require repeatedly evaluating candidate ranks under time-varying CSI, incurring substantial training and signaling overhead. Second, $\mathcal{P}_1$ is a mixed-integer nonlinear programming problem with a long-term latency constraint and strongly coupled variables, making direct online optimization impractical. We therefore adopt a two-timescale design in which the offline-optimized rank $r$ is selected offline before training, whereas $O^t$ and $b_k^t$ are optimized online during training.


\subsection{offline-optimized Rank Selection}
As discussed in the previous subsection, the offline stage is performed before training to determine the appropriate offline-optimized LoRA rank $r$. Since instantaneous CSI is unavailable at this stage, the channel conditions are approximated using statistical CSI. Moreover, because the future channel-dependent sparsification ratios are unknown, each candidate rank is evaluated under a stationary sparsification ratio $O^t=O^0$. Accordingly, when evaluating the offline objective, $\rho_{\max}=(1-O_{\min})/O_{\min}^2$ is specialized to $(1-O^0)/(O^0)^2$. During training, the sparsification ratio $O^t$ is subsequently adapted online according to the instantaneous CSI. Based on these approximations, the offline-stage problem is formulated as
\begin{subequations}\label{OP:P2}
\begin{align}
    \mathcal{P}_2 &\min_{r,O^0, b_k^0} \gamma^0,\\
    \text{s.t.} & \ \  r \leq r_{\max}, ~r\in\mathbb{Z}^+ \label{OP:P2 rank}\\
    & \ \  D^0 \leq D_{th}, \label{OP:P2 latency}\\
    & \ \  O_{\min} \leq O^0 \leq 1, \label{OP:P2 sparsification}\\
    & \ \ 0\leq b_k^0 \leq 1, \quad\sum\nolimits_{k=1}^{N} b_k^0=1, \label{OP:P2 bandwidth}
\end{align}
\end{subequations}
where $\gamma^0$ is the upper-bound on the initial gap in \textbf{Theorem \ref{th: convergence}}. 

Problem $\mathcal{P}_2$ remains an NP-hard MINLP  problem because of the discrete rank variable and the nonlinear coupling among the optimization variables. To make the problem tractable, we enumerate the feasible integer values of $r$ and, for each candidate rank, solve the resulting continuous optimization problem over $O^0$ and $b_k^0$. The rank yielding the minimum objective value is then selected as the offline-optimized rank for training. For a given $r$, the following lemma provides the optimal sparsification ratio $O^0$.

\begin{lemma}\label{lemma:O^0}
    In Problem $\mathcal{P}_2$, for a given integer rank $r$, the optimal sparsification ratio is
    \begin{align}
     O^0=\min\left(1,\frac{N D_{th}}{AKr}\right), \label{eq:O^o}
    \end{align}
    where $A=\sum_{k=1}^{N}\frac{v(d+\ell)}{B\log_2(1+|h_k|^2/\sigma^2)}$.
\end{lemma}

\begin{IEEEproof}
For a fixed rank $r$, increasing $O^0$ retains more model parameters and improves the convergence objective, while increasing the uplink delay. Hence, the optimal $O^0$ is the largest feasible value satisfying the latency constraint. Since the round latency is determined by the slowest scheduled client, the bandwidth allocation is chosen to equalize the transmission delays of the scheduled clients at the latency boundary, i.e., $D_k^0=D_{th}$ for all $k\in\mathcal{K}^0$. Using the scheduling and bandwidth-allocation properties, the resulting latency condition can be written as
\begin{align}
    D_{th}\!=\!\mathbb{E}\left[\sum_{k\in\mathcal{K}^0}\frac{vO^0r(d+\ell)}{B\log_2(1+\frac{|h_k|^2}{\sigma^2})}\right]
    =\frac{K}{N}\sum_{k=1}^{N}\frac{vO^0r(d+\ell)}{B\log_2(1+\frac{|h_k|^2}{\sigma^2})}\label{eq: D_{max}}.
\end{align}
Solving \eqref{eq: D_{max}} for $O^0$ and enforcing $O^0\leq1$ yields \eqref{eq:O^o}
\end{IEEEproof}
Using $\textbf{Lemma \ref{lemma:O^0}}$, the offline stage evaluates each feasible candidate rank and selects the offline-optimized rank that minimizes the corresponding convergence upper bound. Federated fine-tuning is then performed using this fixed offline-optimized rank. At each training iteration, the online stage adapts the sparsification ratio $O^t$ and bandwidth allocation $b_k^t$ according to the instantaneous CSI.

\subsection{Online Sparsification and Bandwidth Control}
In this subsection, we optimize the remaining variables for the offline-optimized LoRA rank selected in the offline stage. Specifically, at each iteration $t$, we dynamically adapt the sparsification ratio $O^t$ and bandwidth allocation $b_k^t$ according to the instantaneous channel conditions. Because these variables are coupled through the uplink latency and must satisfy the long-term latency constraint in \eqref{OP:P1 latency}, we employ Lyapunov optimization to enable online control. To handle the time-average latency constraint, we introduce a virtual queue $Q^t$ that evolves as
\begin{align}
    Q^{t+1}=\max(Q^t+D^t-D_{th},0),
\end{align}
with $Q^0=0$. The virtual queue accumulates excess latency relative to the target $D_{th}$, and its stability ensures satisfaction of the long-term latency constraint. We define the Lyapunov function as $L(Q^t)=\frac{1}{2}(Q^t)^2$ and the corresponding conditional Lyapunov drift as
\begin{align}
    \Delta(Q^t)=\mathbb{E}[L(Q^{t+1})-L(Q^t)|Q^t],
\end{align}
where the expectation is taken over the random system state, i.e. channel state. 

To balance latency-queue stability and convergence performance, we consider the drift-plus-penalty function
\begin{align}\label{eq: drift plus}
    \Delta(Q^t)+V\mathbb{E}\left[\gamma^t|Q^t\right],
\end{align}
where $V>0$ controls the tradeoff between convergence performance and queue stability. A larger $V$ places greater emphasis on minimizing the convergence upper bound.

To obtain a tractable per-iteration objective, we first upper bound the Lyapunov drift. Using the virtual-queue update,
\begin{align} 
&\Delta(Q^t) \nonumber\\
&= \mathbb{E}\left[ \frac{1}{2} (Q^{t+1})^2 - \frac{1}{2} (Q^t)^2 \Big| Q^t \right] \\
&= \mathbb{E}\left[ \frac{1}{2} \left( \max\left\{ Q^t + D^t - D_{th}, 0 \right\} \right)^2 - \frac{1}{2} (Q^t)^2 \Big| Q^t \right]  \\
&\leq \mathbb{E}\left[ \frac{1}{2} \left( Q^t + D^t - D_{th} \right)^2 - \frac{1}{2} (Q^t)^2 \Big| Q^t \right] \label{ineq:Delta_Q}\\
&= Q^t \mathbb{E}\left[ D^t - D_{th} | Q^t \right] + \frac{1}{2} \mathbb{E}\left[ \left( D^t - D_{th} \right)^2 | Q^t \right]\\ 
&\leq \mathcal{V}+ Q^t \left( \mathbb{E}\left[ D^t | Q^t \right] - D_{th} \right), \label{ineq:Delta_Q_2}
\end{align}
where $\mathcal{V}=\tfrac{1}{2}\max\!\left\{D_{th}^2,(\bar
D-D_{th})^2\right\}$ and $\bar{D}$ denotes an upper bound on $D^t$. Such an upper bound can be determined from the minimum feasible bandwidth and worst-case channel conditions. Inequality \eqref{ineq:Delta_Q}  follows from $(\max{x,0})^2\leq x^2$, while \eqref{ineq:Delta_Q_2} follows from $0\leq D^t\leq\bar D$.

Substituting \eqref{ineq:Delta_Q_2} into the drift-plus-penalty function yields an upper bound that can be minimized at each iteration. At the beginning of iteration $t$, the current queue $Q^t$ and instantaneous CSI are known at the server. Therefore, conditioned on the observed system state, both $D^t$ and $\gamma^t$ become deterministic functions of the decision variables. After removing terms independent of $O^t$ and $b_k^t$, the per-iteration optimization problem becomes
\begin{subequations}
\begin{align}
\mathcal{P}_3: &\min_{O^t, b_k^t}  Q^t D^t + V \gamma^t  \\
\text{s.t.} & \ \  O_{\min} \leq O^t \leq 1,  \\ 
& \ \ 0 \leq b_k^t \leq 1, \quad \sum\nolimits_{k=1}^N b_k^t = 1, 
\end{align} 
\end{subequations}
where $\gamma^t$ denotes the iteration-dependent component of the convergence upper bound in \textbf{Theorem \ref{th: convergence}}.

We first eliminate the bandwidth-allocation variables. For a fixed $O^t$, minimizing the maximum uplink delay requires equalizing the transmission delays among the scheduled clients. Define
$R_k^t=\log_2\left(1+\frac{|h_k^t|^2}{\sigma^2}\right)$ and
$A^t=\sum_{j\in\mathcal{K}^t}\frac{1}{R_j^t}$.
The optimal bandwidth allocation is then
\begin{align}\label{eq:b_opt}
b_k^{t}=\frac{1/R_k^t}{\sum_{j\in\mathcal K^t}1/R_j^t}=\frac{1}{A^tR_k^t},\qquad k\in\mathcal K^t,
\end{align} with $b_k^t=0$ for $k\notin\mathcal{K}^t$. Under this allocation, all scheduled clients experience the same transmission delay,
\begin{align}
D^{t}(O^t)=\frac{vO^tr(d+\ell)}{B}A^t
\end{align}

Substituting this expression into $\mathcal{P}3$ reduces the online optimization to the scalar problem
\begin{align}
\mathcal{P}_4: &\min_{O_{\min}\leq O^t\leq1}\quad Q^t\frac{vr(d+\ell)A^t}{B}\,O^t+V\gamma^t.\label{eq:scalar_online_problem}
\end{align}
For fixed $r$, $\gamma^t$ depends on $O^t$ through the sparsification term $\rho^t=\frac{1-O^t}{O^t}$. Let
\begin{align}
\kappa=\frac{256N}{K}r^2S^2W^4\big(3+16\rho_{\max}\big).
\end{align}
Ignoring terms independent of $O^t$, the objective of $\mathcal{P}4$ is proportional to
\begin{align}
Q^t\frac{vr(d+\ell)A^t}{B}O^t+V\kappa\frac{1-O^t}{O^t},
\end{align}
which is convex in $O^t$. For $Q^t>0$, the first-order optimality condition yields
\begin{align}
O^{t}=\min\left\{1,\,\max\left\{O_{\min},\,
\sqrt{\frac{V\kappa B}{Q^t vr(d+\ell)A^t}}\right\}\right\},
\label{eq:O_opt}
\end{align}
When $Q^t=0$, the latency term vanishes and the convergence objective is minimized by $O^t=1$.

At each iteration, the server observes the instantaneous CSI and queue state, computes $O^t$ and $b_k^t$ using \eqref{eq:O_opt} and \eqref{eq:b_opt}, and communicates these decisions to the scheduled clients. The online stage therefore adapts the transmitted payload and bandwidth to the instantaneous channel conditions within the LoRA structure fixed by the offline stage, complementing the slow-timescale offline-optimized-rank selection with fast-timescale communication control.

\section{Numerical Results}\label{sec: numerical results}
In this section, we present simulation results to evaluate the effectiveness of the proposed framework. All experiments were conducted using Python 3.8 on an Ubuntu server equipped with NVIDIA GeForce RTX 3090 GPUs.

We evaluate the proposed method on the CIFAR-100 dataset using a pre-trained ViT-Base model. ViT-Base is a transformer-based architecture designed for image recognition, and LoRA modules are inserted into Blocks 0, 3, 6, and 9. Training is performed using the Adam optimizer with a learning rate of $10^{-4}$ and a weight decay of $10^{-4}$. The local batch size and number of local epochs are set to 8 and 1, respectively.

Following the wireless channel model described in Section~\ref{sec: system model}, $N=100$ clients are placed at distances uniformly drawn from $[100,500]$ m from the server, with path-loss exponent $\chi=3$. Each uplink experiences Rayleigh block fading, where the small-scale fading coefficient $z_k^t\sim\mathcal{CN}(0,1)$ remains constant within each communication round and varies independently across rounds. At each round, $K=10$ clients are scheduled and share a total bandwidth of $B=10$ MHz through FDMA according to the allocation in \eqref{eq:b_opt}. Each client uploads $vO^tr(d+\ell)$ bits of sparsified LoRA parameters, where $v=16$ bits per element. Under the latency threshold $D_{th}=10$ s, the offline stage selects the offline-optimized rank as $r^\ast=32$.

For performance comparison, we consider the following baseline schemes:
\begin{itemize}
\item \textit{\textbf{FLASC} \cite{kuo2024federated}:} Each client applies magnitude-based sparsification to the LoRA matrices by retaining the top-$q$ entries with the largest absolute values and setting the remaining entries to zero.

\item \textit{\textbf{RLoRA} \cite{stich2018sparsified}:} Each client randomly selects entries of the LoRA matrices for transmission, regardless of their magnitudes or structural positions.

\item \textit{\textbf{HetLoRA} \cite{cho2023heterogeneous}:} Each client performs structured sparsification along the rank dimension. Rank components with smaller indices are retained, while those with larger indices are masked according to the target sparsification ratio. The number of retained rank indices is determined by the architectural ordering and remains fixed throughout training.

\item \textit{\textbf{FedLoDrop} \cite{xie2026fedlodrop}:} Each client performs structured sparsification through random row and column dropout. At each iteration, Bernoulli masks are applied to the $d$ rows of $\theta_{B,k}$ and the $\ell$ columns of $\theta_{A,k}$. The number of retained rank indices, corresponding to the selected row and column vectors along the rank dimension, is determined by the target sparsification ratio.

\item \textit{\textbf{FixLoRA}:} Each client transmits entries according to a fixed random sparsity pattern. The pattern is generated before training with a density determined by the target sparsification ratio, shared across clients through a common random seed, and kept unchanged throughout all communication rounds.

\item \textit{\textbf{IRLoRA}:} Each client performs structured sparsification along the rank dimension using the importance metric of the proposed scheme. At each iteration, importance scores are computed from the current local LoRA matrices, and the rank indices with the largest scores are retained in full. The number of retained rank indices is determined by the target sparsification ratio. Unlike \textbf{SOFT}, no element-wise sparsification is applied.
\end{itemize}

\begin{figure}
    \centering
    \includegraphics[width=0.32\textwidth]{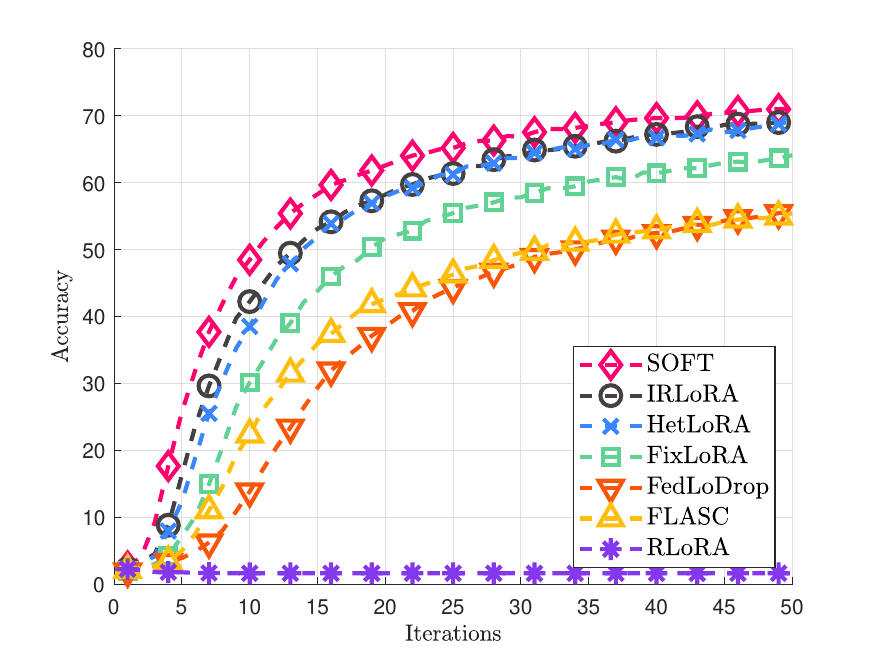}
    \caption{Test accuracy of CIFAR-100 dataset with rank $r=4$.}
    \label{fig: OLoRA_r4} 
\end{figure}
\begin{figure*}[!t]
    \centering
    \setlength{\subfigcapskip}{-2mm}
    \subfigure[Test accuracy]{\includegraphics[width=0.32\linewidth]{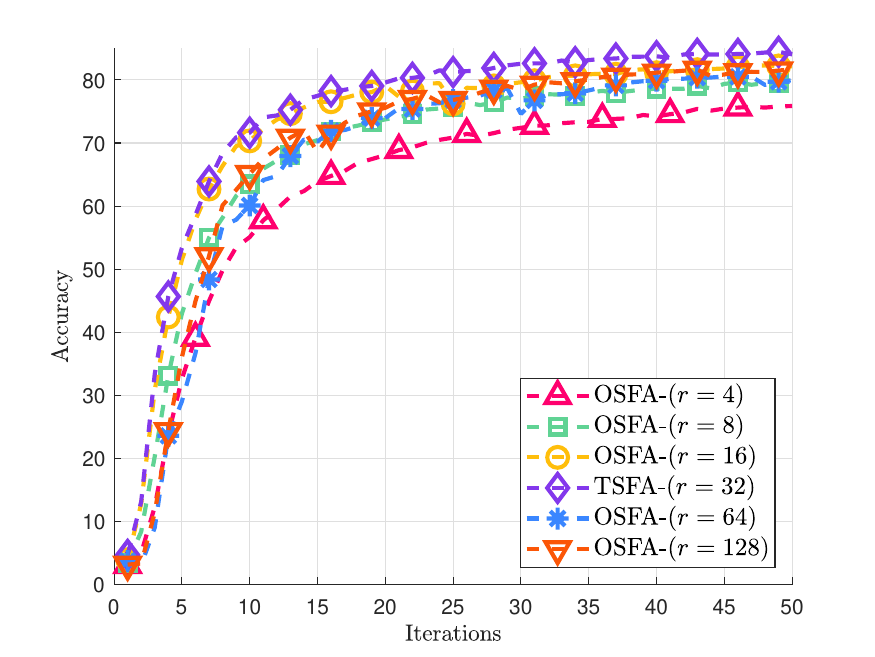}
    \label{fig: two_accuracy}}
    \subfigure[Total delay]{\includegraphics[width=0.32\linewidth]{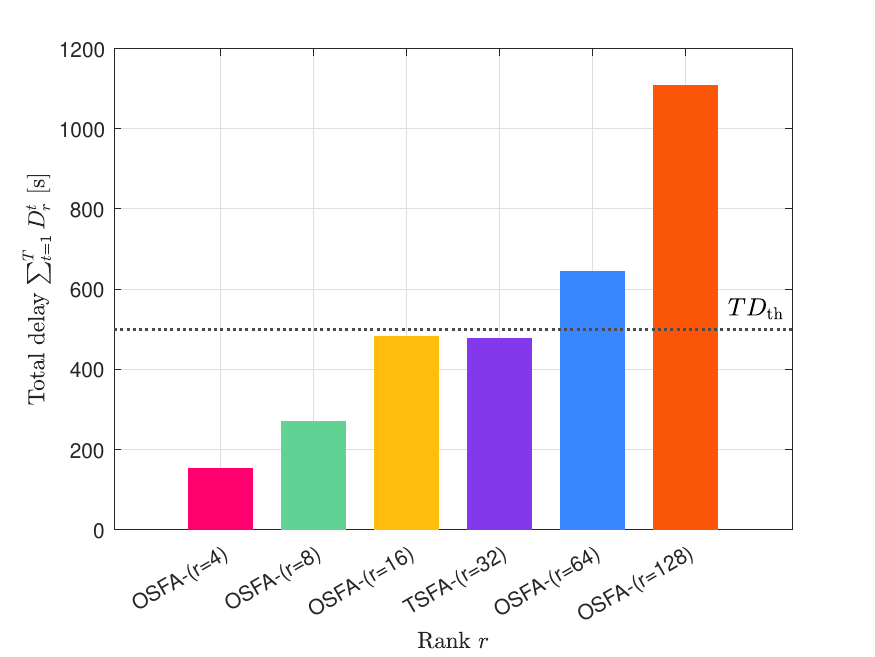}
    \label{fig: total_delay}}
    \subfigure[Average sparsification ratio]{\includegraphics[width=0.32\linewidth]{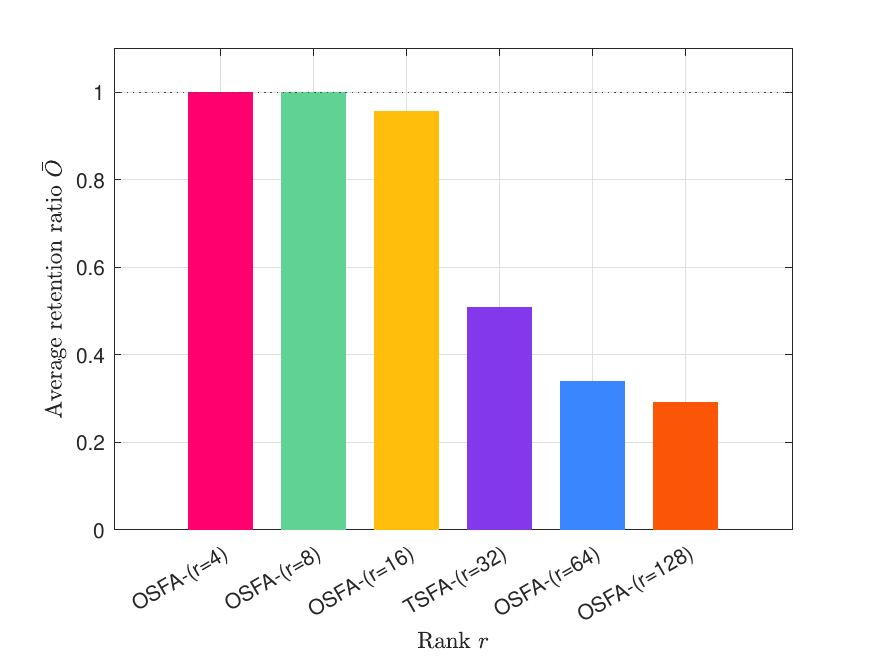}
    \label{fig: retained_ratio}}
     \caption{Performance of \textbf{OSFA} and \textbf{TSFA}: (a) test accuracy on CIFAR-100, (b) total
latency over the training, and (c) average sparsification ratio.}
\label{fig: two-stage}
\end{figure*}

\subsection{Effects of \textbf{SOFT}}\label{sec: S-OLoRA}

In this subsection, we evaluate the effectiveness of \textbf{SOFT} by comparing it with the baseline methods. Fig.~\ref{fig: OLoRA_r4} shows the test accuracy on CIFAR-100 with LoRA rank $r=4$. For a fair comparison, the sparsification ratio is fixed at $0.5$ for all methods. \textbf{SOFT} consistently achieves the highest accuracy, demonstrating the benefit of dynamically allocating the sparsification budget according to rank-component importance while retaining the most significant entries within each component.

\textbf{IRLoRA} underperforms \textbf{SOFT} because it performs sparsification at the rank-component level. For each retained rank index, all entries in the corresponding row and column of the two LoRA matrices are transmitted, even if some make negligible contributions. Conversely, rank components whose importance scores fall below the selection threshold are discarded entirely.
\textbf{HetLoRA}, which consistently retains components with lower rank indices, underperforms \textbf{SOFT} because its selection does not adapt to the evolving importance of rank components during training. \textbf{FixLoRA} uses the same sparsity pattern at every iteration, allowing the retained entries to be trained consistently; however, because the pattern is randomly determined before training, it does not reflect parameter importance. \textbf{FLASC} and \textbf{FedLoDrop} achieve similar accuracy, both below \textbf{FixLoRA}. \textbf{FLASC} selects entries solely by magnitude, ignoring the multiplicative coupling between the two LoRA matrices and potentially discarding entries that are small individually but influential through the matrix product. \textbf{FedLoDrop} randomly masks different rows and columns at each iteration, so its retained structure neither reflects component importance nor remains consistent across training. Finally, \textbf{RLoRA} fails to achieve meaningful learning, with its test accuracy remaining nearly unchanged throughout training.

\subsection{Effects of offline-optimized Rank Selection}\label{sec: S-Offline}
In Section~\ref{sec: WFLoRA}, we introduced a two-stage framework consisting of an offline stage for offline-optimized rank selection and an online stage for dynamic adaptation during training. To isolate the contribution of the offline stage, we compare \textbf{TSFA} with a one-stage federated algorithm (\textbf{OSFA}). \textbf{OSFA} omits the offline optimization and instead fixes the offline-optimized rank at a prescribed value, consistent with the common practice of treating the rank as a manually tuned hyperparameter \cite{zhang2024towards,sun2024improving,babakniya2023,yi2025fedalora}. Both algorithms use the same online control for the sparsification ratio and bandwidth allocation; hence, any performance difference arises solely from the offline-optimized rank selection. Under $\beta=1.0$, $D_{th}=10$s, and $V=10^{-6}$, the offline stage of \textbf{TSFA} selects $r^\ast=32$. Fig.~\ref{fig: two_accuracy} compares the test accuracy of \textbf{TSFA} at $r^\ast$ with that of \textbf{OSFA} over alternative offline-optimized ranks, with all configurations trained for the same number of iterations. \textbf{TSFA} achieves the highest accuracy, validating the offline rank selection. For $r<r^\ast$, the limited model capacity leads to under-parameterization, degrading accuracy while leaving part of the latency budget unused. For $r>r^\ast$, the larger update requires more aggressive sparsification, causing informative model updates to be discarded. As shown below, these oversized offline-optimized ranks also violate the latency constraint despite the stronger compression. Thus, an excessively large offline-optimized rank is detrimental to both learning accuracy and latency feasibility.

We next examine how the latency constraint shapes this behavior. Figs.~\ref{fig: total_delay} and \ref{fig: retained_ratio} show the total latency and average sparsification ratio $\bar{O}$ over the same rank sweep. The total latency constraint $TD_{th}$ is indicated by the horizontal line in Fig.~\ref{fig: total_delay}. For $r<r^\ast$, the payload is sufficiently small that the online stage maintains $\bar{O}$ close to one, resulting in substantial unused latency margin. At $r^\ast$, the total latency approaches the constraint, indicating efficient use of the available communication budget. For $r>r^\ast$, the online stage increasingly reduces $\bar{O}$ to compensate for the larger payload, yet the total latency still exceeds the constraint. These violations therefore arise from an oversized offline-optimized rank rather than insufficient adaptation by the online stage. Overall, the results demonstrate that the offline stage selects a offline-optimized rank that effectively balances model capacity and latency feasibility.


\subsection{Effects of Online Resource Control}
\label{sec: S-CDF}
\begin{figure}
    \centering
    \includegraphics[width=0.32\textwidth]{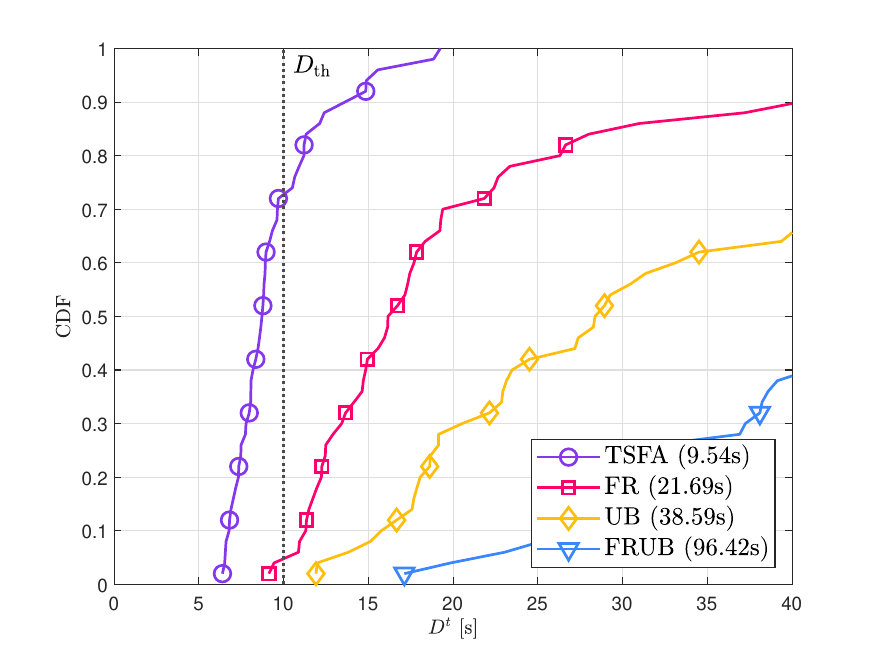}
    \caption{CDF of the $D^t$ under different
resource allocation schemes.}
    \label{fig: cdf}
\end{figure}

Fig.~\ref{fig: cdf} evaluates the effectiveness of the online stage by comparing the per-iteration delay distributions at $r^\ast=32$. All schemes use the same offline-optimized rank selected by the offline stage. Fixed ratio (\textbf{FR}) fixes the sparsification ratio at the offline-stage value $O^0$, uniform bandwidth (\textbf{UB}) retains online sparsification control but allocates bandwidth uniformly, and fixed ratio with uniform bandwidth (\textbf{FRUB}) applies both modifications. The value of each CDF at the vertical line $D_{th}$ represents the fraction of iterations satisfying the per-iteration deadline, while the legend reports the corresponding time-average delay. We emphasize that the constraint in \eqref{OP:P1 latency} is imposed on the time-average delay rather than on every individual iteration. Hence, a scheme can satisfy the constraint even if some iterations exceed $D_{th}$, provided that the average delay remains below the threshold. Such occasional violations under deep fading are therefore allowed by design and compensated for over time.

\textbf{TSFA} exhibits this desired behavior: its delay distribution is concentrated near the threshold with a relatively short tail, while its time-average delay satisfies the constraint. In contrast, the three variants violate the constraint for different reasons. \textbf{FR} fails because the offline-stage ratio $O^0$, determined from statistical channel information, cannot account for instantaneous fading; consequently, deep fades can produce delays substantially larger than those predicted from average channel conditions. \textbf{UB} fails because uniform bandwidth allocation leaves the round delay dominated by the worst-channel scheduled client. Sparsification control alone cannot fully eliminate this bottleneck because $O^t$ is lower bounded by $O_{\min}$, resulting in a time-average delay even larger than that of \textbf{FR}. \textbf{FRUB} combines both limitations and therefore yields the largest time-average delay. These results confirm that the two online controls address complementary effects: sparsification adaptation compensates for channel-induced payload constraints, while bandwidth allocation mitigates the slowest-client bottleneck.

\begin{figure}
    \centering
    \includegraphics[width=0.32\textwidth]{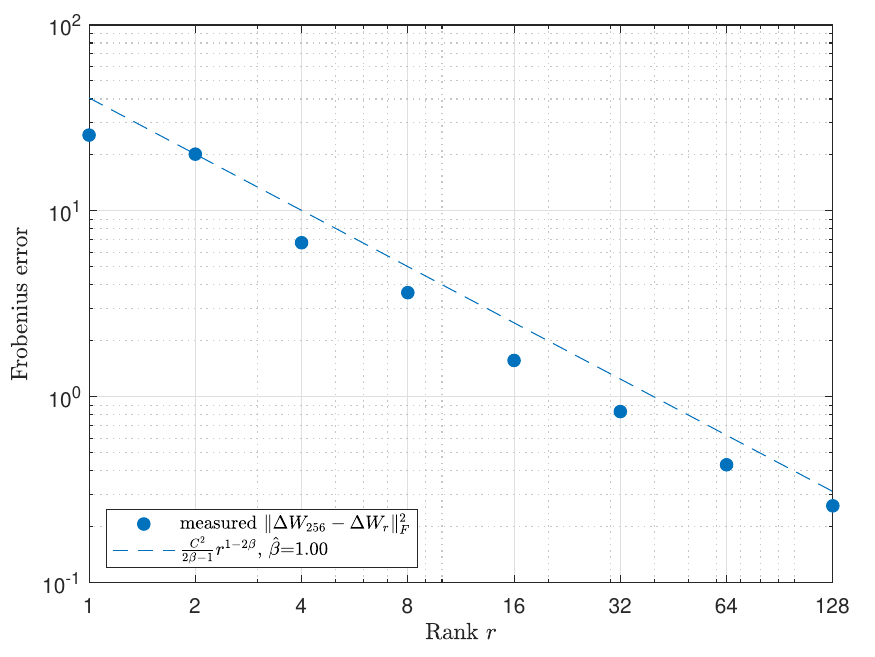}
    \vspace{-3mm}
    \caption{Frobenius error of the trained rank-$r$ model (log--log scale).}
    \label{fig: as2}
    \vspace{-5mm}
\end{figure}

\vspace{-2mm}
\subsection{Empirical Validation of Assumption~\ref{as: lora error}}
\label{sec: validation}
We validate \textbf{Assumption~\ref{as: lora error}} through a matched rank sweep. A LoRA model with rank $r=256$ is used as a proxy for $\theta_o$, and rank-$r$ models with $r\in{1,2,4,8,16,32,64,128}$ are trained on CIFAR-100 under the same setting. For each $r$, we construct the effective weight update $\Delta W_r=\theta_B\theta_A$ for every adapted layer and measure the approximation error as  $\sum_{\ell}\|\Delta W_{256}^{(\ell)}-\Delta W_r^{(\ell)}\|_F^2$, summed over all adapted layers. Fig.~\ref{fig: as2} plots the measured error versus $r$ on logarithmic axes, together with the bound in \eqref{eq:lora_error_bound}, where $C$ is chosen as the smallest constant such that the bound covers all measured points. The approximation error closely follows a single power-law decay, with the fitted exponent $\hat\beta=1.00$ comfortably exceeding the threshold $1/2$ required by Assumption~\ref{as: lora error}. Moreover, the bound remains within a factor of $1.6$ of the measured error across all ranks. Similar behavior is observed at the individual-layer level, where every adapted layer exhibits a fitted exponent $\hat\beta_\ell\in[0.88,1.11]$, confirming that the aggregate decay is not dominated by any single layer.

\begin{figure}
    \centering
    \includegraphics[width=0.32\textwidth]{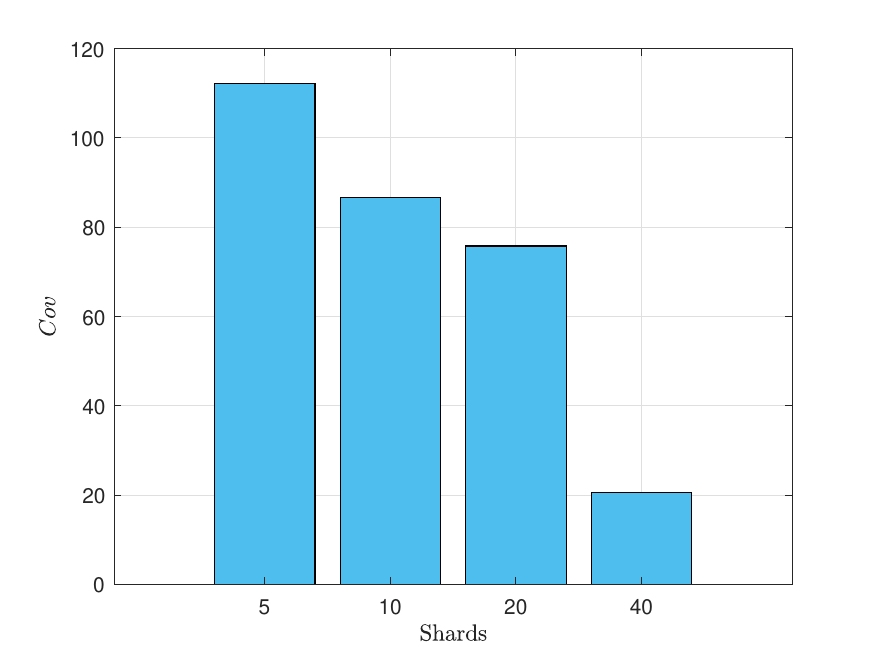}
    \vspace{-3mm}
    \caption{Frobenius norm of covariance.}
    \label{fig: non_iid_bar}
    \vspace{-5mm}
\end{figure}
\begin{figure}
    \centering
    \includegraphics[width=0.32\textwidth]{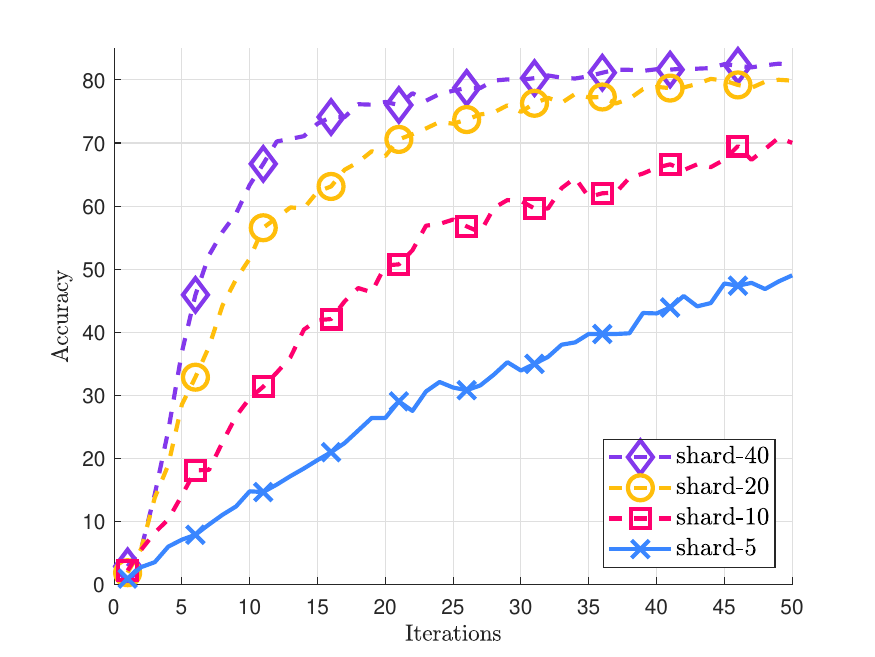}
    \vspace{-3mm}
    \caption{Test accuracy of CIFAR-100 under different number of shards.}
    \label{fig: non_iid_graph}
    \vspace{-5mm}
\end{figure}
\subsection{Effects of Covariance}\label{sec: cov}

In Section~\ref{sec: WFLoRA}, we introduced the covariance term that arises from the separate transmission and aggregation of the two LoRA matrices. In this subsection, we provide empirical evidence showing how this term varies with the degree of data heterogeneity under non-IID data distributions. To generate heterogeneous client datasets, we adopt a shard-based non-IID partitioning scheme. Specifically, the dataset is first sorted by class label and divided into multiple shards, each containing samples predominantly from a single class. These shards are then randomly assigned to clients, resulting in client datasets biased toward different subsets of classes.

Fig.~\ref{fig: non_iid_bar} shows the evolution of the covariance term during training under different shard sizes, while Fig.~\ref{fig: non_iid_graph} presents the corresponding test accuracy. We set $r=32$ and remove communication constraints to isolate the effects of data heterogeneity. Under the adopted partitioning scheme, increasing the shard size makes the resulting client distributions closer to IID, thereby reducing the discrepancy among locally updated LoRA parameters. Accordingly, the covariance magnitude decreases as the data distributions become more homogeneous, as shown in Fig.~\ref{fig: non_iid_bar}. Fig.~\ref{fig: non_iid_graph} further shows that settings with lower covariance generally achieve higher test accuracy, providing empirical support for the relationship between client heterogeneity, the covariance term, and learning performance. These observations motivate future approaches that mitigate the separate-aggregation discrepancy through improved data allocation, adaptive aggregation, or LoRA-specific training strategies.

\vspace{-5mm}
\section{Conclusion}\label{sec: conclusion}
This paper studied federated LoRA fine-tuning over time-varying wireless networks by distinguishing two timescales of decisions: the \emph{offline-optimized rank}, fixed before training, whereas the sparsification ratio and bandwidth allocation are adapted during training. We proposed \textbf{SOFT}, which promotes near-orthogonality among rank components and uses factor-norm products as low-complexity spectral importance scores to allocate the transmission budget across rank components and select entries within each component. We further developed \textbf{TSFA}, whose offline stage selects a latency-feasible offline-optimized rank using statistical channel information, while its online stage adapts the sparsification ratio and bandwidth allocation using instantaneous channel information. The design is guided by a convergence bound that captures offline-optimized-rank approximation error, partial client participation, sparsification error, and the discrepancy induced by separate aggregation of the two LoRA matrices. Overall, the proposed framework jointly accounts for learning performance and wireless communication constraints, enabling communication-efficient federated LoRA fine-tuning under time-varying channel conditions.

\appendices

\section{Proof of Lemma \ref{lemma:error_bound}}\label{apdx of lemma:error_bound}

Before proceeding with the main proof, we first establish essential inequality. Using Arithmetic and Geometric Mean inequality (AM-GM  inequality), i.e. $\gamma a+\frac{1}{\gamma}b\geq 2\sqrt{ab}$ for $\gamma>0$, $\|x+y\|_F^2$ can be upper-bounded as:
\begin{align}
    \|x+y\|_F^2 &= \|x\|_F^2+\|y\|_F^2+2x^\textsf{T}y\\
    &\leq \|x\|_F^2+\|y\|_F^2 +\gamma\|x\|_F^2+\frac{1}{\gamma}\|y\|_F^2\\
    &=(1+\gamma)\|x\|_F^2+\left(1+\frac{1}{\gamma}\right)\|y\|_F^2.\label{eq: gamma AM-GM}
\end{align}
Now, using the \textbf{Assumption \ref{as: sparsification}} and \eqref{eq: overall memory update}, 
\begin{align}
   \left\|\tilde{m}_k^{t+1}\right\|_F^2&\leq (1-O^t) \mathbb{E}\left[\|\tilde{m}_k^t+\tilde{\theta}_k^t\|_F^2\right]\\
    &\leq(1-O^t)(1+\gamma)\mathbb{E}[\|\tilde{m}_k^t\|_F^2]\nonumber\\
    &\quad+(1-O^t)\left(1+\frac{1}{\gamma}\right)\|\tilde{\theta}_k^t\|_F^2\label{eq: recursive eq},
\end{align}
where the second inequality  follows  from \eqref{eq: gamma AM-GM}. Choosing $\gamma=\frac{O^t}{2(1-O^t)}$ gives
\begin{align}
(1-O^t)(1+\gamma)=1-\frac{O^t}{2},
\end{align}
and
\begin{align}
(1-O^t)\left(1+\frac{1}{\gamma}\right)
=\frac{(1-O^t)(2-O^t)}{O^t}
\leq 2\rho^t,
\end{align}
where $\rho^t=\frac{1-O^t}{O^t}$. This proves \eqref{eq:one_step_memory}, To establish \eqref{eq:uniform_memory}, we instead use the fixed choice
$\gamma=\frac{O_{\min}}{2(1-O_{\min})}$.
Since $O^i\geq O_{\min}$ for all $i\leq t$, we have $1-O^i\leq1-O_{\min}$. Define
\begin{align}
\Gamma=(1-O_{\min})\left(1+\frac{1}{\gamma}\right).
\end{align}
Recursively applying \eqref{eq: recursive eq} then yields
\begin{align}
    &\|\tilde{m}_k^{t+1}\|_F^2\nonumber\\
    &\leq \Gamma\sum\nolimits_{i=0}^t\left[(1-O_{\min})(1+\gamma)\right]^{t-i}\|\tilde{\theta}_k^i\|_F^2\\
    & \leq\Gamma\sum\nolimits_{j=0}^{t}\left[(1-O_{\min})(1+\gamma)\right]^{j}\max_{0\leq i \leq t}\|\tilde{\theta}_k^i\|_F^2 \label{inq: max}\\
    &\leq\Gamma\sum\nolimits_{j=0}^\infty\left[(1-O_{\min})(1+\gamma)\right]^{j}\max_{0\leq i \leq t}\|\tilde{\theta}_k^i\|_F^2\\
    &= \frac{2(1-O_{\min})(2-O_{\min})}{(O_{\min})^2}\max_{0\leq i \leq t}\|\tilde{\theta}_k^i\|_F^2\label{eq: inf}\\
    &\leq\frac{4(1-O_{\min})}{(O_{\min})^2}\max_{0\leq i \leq t}\|\tilde{\theta}_k^i\|_F^2,
\end{align}
where  \eqref{inq: max} follows by  $\|\tilde{\theta}_k^i\|_F^2\leq\max_{0\leq i \leq t}\|\tilde{\theta}_k^i\|_F^2$ and \eqref{eq: inf} is obtained by $\gamma=\frac{O_{\min}}{2(1-O_{\min})}$, which  ensures $(1+\gamma)(1-O_{\min})<1$ for $0<O_{\min}<1$.   Finally, $2-O_{\min}\leq2$ gives the last inequality, completing the proof. 

\section{Proof of Theorem \ref{th: convergence}}\label{apdx of th: convergence}

Let $\theta_o$ be an original model. Then the update of the original and LoRA model at the  $t+1$-th global iteration can be described as:
\begin{align}
    \theta_o^{t+1}&=\sum\nolimits_{k\in\mathcal{N}}p_k\theta_{o,k}^{t+1}\nonumber\\
    &=\theta_o^t-\sum\nolimits_{k\in\mathcal{N}}p_k\eta\sum_{e=0}^{E-1}\nabla F_{o,k}(\theta_{o,k}^{t,e}),\label{eq: theta_o}\\
    \theta_r^{t+1}&=\frac{N}{K}\sum_{k\in\mathcal{K}^{t}}p_k\theta_{r,k}^{t+1}-\frac{N}{K}\sum_{k\in\mathcal{K}^{t}}p_k\mu_k^t-\textit{Cov}(\theta_B^{t+1},\theta_A^{t+1})\nonumber\\
    &=\theta_r^t-\frac{N}{K}\sum_{k\in\mathcal{K}^{t}}p_k\eta\sum_{e=0}^{E-1}\nabla F_{r,k}(\theta_{r,k}^{t,e})-\frac{N}{K}\sum_{k\in\mathcal{K}^{t}}p_k\mu_k^t\nonumber \\
    &\quad\quad- \textit{Cov}(\theta_B^{t+1},\theta_A^{t+1}), \label{eq: theta_r}
\end{align}
where $\mu_k^t = \theta_{B,k}^t\theta_{A,k}^t
-\psi_{B,k}^t\psi_{A,k}^t$ denotes the gap between locally trained update and its sparsified update at iteration $t$.

Using $S$-smoothness, 
\begin{align}\label{eq: 1-round convergence}
    \mathbb{E}[F_o(\theta_o^{t+1})-F_o(\theta_o^t)]&\leq\underbrace{\mathbb{E}[\nabla F_o(\theta_o^t)^\textsf{T}(\theta_o^{t+1}-\theta_o^t)]}_{(a)}\nonumber\\
    &+\frac{S}{2}\underbrace{\mathbb{E}[\|\theta_o^{t+1}-\theta_o^t\|_F^2]}_{(b)}.
\end{align}
For $(a)$, using \eqref{eq: theta_o}, we can obtain
\begin{align}
    (a)&=-\eta\sum_{e=0}^{E-1}\mathbb{E}\left[\nabla F_o(\theta_o^t)^\textsf{T}\sum_{k\in \mathcal{N}}p_k\nabla F_{o,k}(\theta_o^{t,e})\right]\\
    &=-\frac{\eta}{2}\underbrace{\sum\nolimits_{e=0}^{E-1}\mathbb{E}\left[\left\|\nabla F_o(\theta_o^t)\right\|_F^2\right]}_{(a1)}\nonumber\\
    &-\frac{\eta}{2}\underbrace{\sum\nolimits_{e=0}^{E-1}\mathbb{E}\left[\left\|\sum\nolimits_{k\in\mathcal{N}}p_k\nabla F_{o,k}(\theta_{o,k}^{t,e})\right\|_F^2\right]}_{(a2)}\nonumber\\
    &+\frac{\eta}{2}\underbrace{\sum_{e=0}^{E-1}\mathbb{E}\left[\left\|\nabla F_o(\theta_o^t)-\sum_{k\in\mathcal{N}}p_k\nabla F_{o,k}(\theta_{o,k}^{t,e})\right\|_F^2\right]}_{(a3)},
\end{align}
where the second equality is due to the property $-\mathbf{a}^\textsf{T}\mathbf{b} = \frac{1}{2}(\!-\|\mathbf{a}\|_2^2 \!-\! \|\mathbf{b}\|_2^2 \!+\! \|\mathbf{a \!-\! b}\|_2^2)$.  Next, we use the following lemma:
\begin{lemma}\label{lemma: bound of (a2)}
Under \textbf{Assumptions \ref{as: sparsification}, \ref{as: lora error}, \ref{as: covariance},
\ref{as: norm boundedness}}, and
\textbf{Lemma~\ref{lemma:error_bound}}, the inequality below is
satisfied:
    \begin{align}
        -&\sum_{e=0}^{E-1}\mathbb{E}\left[\left\|\sum_{k\in\mathcal{N}}p_k\nabla F_{o,k}(\theta_{o,k}^{t,e})\right\|_F^2\right]\leq \!-\frac{1}{2}\mathbb{E}[\|\nabla F_r(\theta_{r}^{t})\|_F^2]\nonumber\\
        &+\! \frac{2S^2C^2}{2\beta-1}\,r^{1-2\beta} +\frac{16(N\!-\!K)}{K(N\!-\!1)}\eta^2S^2E^2G^2+16S^2\phi rW^2\nonumber\\
        &+\frac{128N r^2S^2W^4}{K}(\rho^t\!+\!4\rho_{\max})\!(3+16\rho_{\max}).
    \end{align}
\end{lemma}
\begin{IEEEproof}
    See \emph{Appendix \ref{apdx of lemma: bound of (a2)}}.
\end{IEEEproof}
Using \textbf{Lemma \ref{lemma: bound of (a2)}}, we can derive an upper-bound on $(a2)$. Next, using the $S$-smoothness, 
Jensen's inequality and the properties $\theta_o^t=\theta_{o,k}^{t,0}$ and  $\sum_{k\in\mathcal{N}}p_k=1$,   we can obtain an upper-bound on $(a3)$ as:  
\begin{align}
    (a3)&\leq S^2 \sum_{k\in\mathcal{N}}p_k\sum_{e=0}^{E-1}\mathbb{E}\left[\left\|\theta_{o,k}^{t,0}-\theta_{o,k}^{t,e}\right\|_F^2\right] \label{eq:a3_1}\\
    &=S^2 \sum_{k\in\mathcal{N}}p_k\sum_{e=0}^{E-1}\mathbb{E}\left[\left\|\sum_{i=0}^{e-1}\eta \nabla F_{o,k}(\theta_{o,k}^{t,e})\right\|_F^2\right] \label{eq:a3_2} \\
    &\leq S^2\eta^2\sum_{k\in\mathcal{N}}p_k\sum_{e=0}^{E-1}e^2G_k^2 \label{eq:a3_3}\\
    &= \frac{E(E-1)(2E-1)S^2\eta^2}{6}\sum_{k\in\mathcal{N}}p_kG_k^2 \label{eq:a3_4}\\
    &\leq \frac{E(E-1)(2E-1)S^2\eta^2}{6}G^2, \label{eq:a3_5}
\end{align}
where the first equality comes from \eqref{eq: theta_o}; the first inequality is due to Jensen's inequality and \textbf{Assumption \ref{as: norm boundedness}}; the last inequality is by $G^2=\max_kG_k^2$  and $\sum_{k\in\mathcal{N}}p_k=1$.  

Similarly, using \eqref{eq: theta_o} and  Jensen's inequality, $(b)$  in \eqref{eq: 1-round convergence} can be upper-bounded as:  
\begin{align}
    (b)&\leq \eta^2 \mathbb{E}\left[\sum_{k\in\mathcal{N}}p_k\sum_{e=0}^{E-1}\left\|\nabla F_{o,k}^{t,e}(\theta_o^{t,e})\right\|_F^2\right]\\
    &\leq\eta^2 E^2\sum_{k\in\mathcal{N}}p_kG_k^2,\\
    &\leq\eta^2 E^2G^2,
\end{align}
where  the second inequality comes from \textbf{Assumption \ref{as: norm boundedness}}. The last inequality is due to $G^2=\max_k{G_k^2}$ and $\sum_{k\in\mathcal{N}}p_k=1$. 

Finally, applying the above inequalities to \eqref{eq: 1-round convergence} and using $(a1)>0$ always holds, we can obtain
\begin{align}
    \mathbb{E}&[F_o(\theta_o^{t+1})-F_o(\theta_o^t)]\leq-\frac{\eta}{4}\mathbb{E}[\|\nabla F_r(\theta_{r}^{t})\|_F^2]+\frac{S}{2}\eta^2E^2G^2\nonumber\\
    &+ \frac{\eta S^2C^2}{2\beta-1}\,r^{1-2\beta}+ 8\eta S^2 \phi rW^2+ \frac{8\eta^3(N-K)}{K(N-1)}S^2E^2G^2
    \nonumber \\
    &+ \frac{64\,\eta N r^2S^2W^4}{K}
\big(\rho^t+4\rho_{\max}\big)\big(3+16\rho_{\max}\big)\nonumber\\
    &+ \frac{\eta^3E(E-1)(2E-1)}{12}S^2G^2.
\end{align}
Averaging the above inequality over iterations from $0$ to $T-1$ and
using the property that $\theta_o^0=\theta_r^0$ since all LoRA
modules are initialized with $\mathbf{0}$ at the initial stage. 
Substituting this into the averaged inequality, we finally obtain \textbf{Theorem \ref{th: convergence}}. 
\vspace{-3mm}
\section {Proof of Lemma \ref{lemma: bound of (a2)}\label{apdx of lemma: bound of (a2)}}
First, we choose only $e=0$ from the summation of positive norm values, then we obtain 
\begin{align}
    &\sum_{e=0}^{E-1}\mathbb{E}\left[\left\|\sum_{k\in\mathcal{N}}p_k\nabla F_{o,k}(\theta_{o,k}^{t,e})\right\|_F^2\right]\\ 
    &\geq \mathbb{E}\left[\left\|\sum_{k\in\mathcal{N}}p_k\nabla F_{o,k}(\theta_{o,k}^{t,0})\right\|_F^2\right]
    =\mathbb{E}[\|\nabla F_o(\theta_{o}^{t})\|_F^2],\label{eq:proof_lemma3}
\end{align} where the equality is by the initialization of the local models, $\theta_{o,k}^{t,0}=\theta_o^t$ for every client $k$. Next, we define $\nabla F_r(\theta_r^t)$ by the gradient of $F_o$ evaluated at the rank-$r$ model, i.e., $\nabla F_r(\theta_r^t)=\nabla F_o(\theta_r^t)$.  Using a simple property, $-\|a\|^2\leq-\frac{1}{2}\|b\|^2+\|a-b\|^2$, we have 
\begin{align}
    -\mathbb{E}[\|\nabla &F_o(\theta_o^t)\|_F^2]\leq-\frac{1}{2}\mathbb{E}[\|\nabla F_r(\theta_r^t)\|_F^2]\nonumber\\
    &+\underbrace{\mathbb{E}[\|\nabla F_o(\theta_o^t)-\nabla F_r(\theta_r^t)\|_F^2]}_{(c)}.
\end{align}
Then,  from  $S$-smoothness, $(c)$ is upper bounded as 
\begin{align}
    (c) &\leq S^2 \mathbb{E}[\|\theta_{o}^t-\theta_r^t\|_F^2]\\
    &\leq 2S^2 \underbrace{\mathbb{E}\left[\left\|\theta_o^t-\hat{\theta}_r^t\right\|_F^2\right]}_{(c1)} + 2S^2 \underbrace{\mathbb{E}\left[\left\|\hat{\theta}_r^t-\theta_r^t\right\|_F^2\right]}_{(c2)}
\end{align}
where the second inequality is due to $\|a-b\|_F^2\leq2\|a\|_F^2+2\|b\|_F^2$ and $\hat{\theta}_r^t=\sum_{k\in\mathcal{N}}p_k\theta_{B,k}^t\theta_{A,k}^t$. 
For $(c1)$,
\begin{align}
    (c1)&\leq\mathbb{E}\left[\sum_{k\in\mathcal{N}}p_k\left\|\theta_{o,k}^{t+1}-\hat{\theta}_{r,k}^{t+1}\right\|_F^2\right]
    \leq \frac{C^2}{2\beta-1}\,r^{1-2\beta},
\end{align}
where the first inequality comes from Jensen's inequality and the
second from \textbf{Assumption \ref{as: lora error}}. Next, plugging
\eqref{eq: theta_r}, $(c2)$ can be expressed as
\begin{align}
    (c2)&\leq2\underbrace{\mathbb{E}\left[\left\|\sum_{k\in\mathcal{N}}p_k\theta_{r,k}^t-\frac{N}{K}\sum_{k\in\mathcal{K}^t}p_k\theta_{r,k}^t\right\|_F^2\right]}_{(c21)}\nonumber\\
    &+2\underbrace{\mathbb{E}\left[\left\|\frac{N}{K}\sum_{k\in\mathcal{K}^t}p_k\mu_k^t-\textit{Cov}(\theta_B^{t+1},\theta_A^{t+1})\right\|_F^2\right]}_{(c22)}.
\end{align}
We follow the same steps as in \textbf{Lemma 5} of
\cite{li2019convergence} to derive the upper-bound on $(c21)$, which
utilize the probabilistic sampling properties
and $\mathbb{E}[\|x-\mathbb{E}[x]\|_F^2]\leq\mathbb{E}[\|x\|_F^2]$.
\begin{align}
    (c21)\leq\frac{4(N-K)}{K(N-1)}\eta^2E^2G^2,
\end{align}
where $G^2=\max_kG_k^2$. 

Before bounding $(c22)$, we first bound
$\mu_k^t=\theta_{B,k}^t\theta_{A,k}^t-\psi_{B,k}^t\psi_{A,k}^t$.
Adding and subtracting $\theta_{B,k}^t\psi_{A,k}^t$, followed by
substituting \eqref{eq: memory update of B}--\eqref{eq: memory update of A},
yields
\begin{align}
\mu_k^t=\theta_{B,k}^t\big(m_{A,k}^{t+1}-m_{A,k}^t\big)
+\big(m_{B,k}^{t+1}-m_{B,k}^t\big)\psi_{A,k}^t.
\end{align}
For either $C\in\{A,B\}$, using $\|a-b\|_F^2\leq2\|a\|_F^2+2\|b\|_F^2$ and
$\|m_{C,k}^t\|_F\leq\|\tilde m_k^t\|_F$,
\begin{align}
\|m_{C,k}^{t+1}-m_{C,k}^t\|_F^2
&\leq 2\|\tilde m_{k}^{t+1}\|_F^2+2\|\tilde m_{k}^t\|_F^2\\
&\leq (4-O^t)\|\tilde m_{k}^{t}\|_F^2
+4\rho^t\|\tilde\theta^t_k\|_F^2\\
&\leq 8rW^2(\rho^t+4\rho_{\max}),
\end{align}
where the second inequality follows from \textbf{Lemma \ref{lemma:error_bound}} and the last inequality follows from \textbf{Assumption \ref{as: norm boundedness}} and \textbf{Lemma \ref{lemma:error_bound}}. Since $\psi_{A,k}^t$ is obtained by retaining a subset of the entries of $m_{A,k}^t+\theta_{A,k}^t$, we also have
\begin{align}
\|\psi_{A,k}^t\|_F^2&\leq\|m_{A,k}^t+\theta_{A,k}^t\|_F^2\nonumber\\
&\leq2\|m_{A,k}^t\|_F^2+2\|\theta_{A,k}^t\|_F^2\nonumber\\
&\leq2rW^2\big(1+8\rho_{\max}\big),\label{eq:psi_bound}
\end{align}
where the second inequality is due to $\|\mathbf{a}+\mathbf{b}\|_F^2\leq2\|\mathbf{a}\|_F^2
+2\|\mathbf{b}\|_F^2$ and the last inequality follows from \textbf{Assumption \ref{as: norm boundedness}} and \textbf{Lemma \ref{lemma:error_bound}}. Combining these bounds with
$\|\mathbf{a}+\mathbf{b}\|_F^2\leq2\|\mathbf{a}\|_F^2
+2\|\mathbf{b}\|_F^2$ and the submultiplicativity of the Frobenius
norm yields
\begin{align}
\|\mu_k^t\|_F^2 &\leq2\|\theta_{B,k}^t\|_F^2\|m_{A,k}^{t+1}-m_{A,k}^t\|_F^2\nonumber\\
&\quad+2\|m_{B,k}^{t+1}-m_{B,k}^t\|_F^2\|\psi_{A,k}^t\|_F^2\nonumber\\
&\leq16r^2W^4\big(\rho^{t}+4\rho_{\max}\big)\big(3+16\rho_{\max}\big).\label{eq:mu_bound}
\end{align}
Using $\|a-b\|^2\leq2\|a\|^2+2\|b\|^2$, $(c22)$ can be bounded as
\begin{align}
    (c22)&\leq
2\mathbb{E}\left[\Big\|\frac{N}{K}\sum_{k\in\mathcal{K}^t}
p_k\mu_k^t\Big\|_F^2\right]
+2\mathbb{E}\left[\big\|\textit{Cov}(\theta_B^{t+1},
\theta_A^{t+1})\big\|_F^2\right]\\
&\leq\frac{2N}{K}\mathbb{E}\Bigg[\frac{N}{K}\sum_{k\in\mathcal{K}^t}
p_k\|\mu_k^t\|_F^2\Bigg]+4\phi rW^2\\
&\leq\frac{32N}{K}r^2W^4(\rho^t\!+\!4\rho_{\max})
(3\!+\!16\rho_{\max})\!+\!4\phi rW^2\!,\!
\end{align}
where the second inequality comes from Jensen's inequality and \textbf{Assumption \ref{as: covariance}}. The last inequality is due to \eqref{eq:mu_bound}.

\ifCLASSOPTIONcaptionsoff
  \newpage
\fi
\bibliographystyle{IEEEtran}
\bibliography{references}

\end{document}